\documentclass[fleqn,10pt]{wlpeerj}
\usepackage{amssymb}
\usepackage{latexsym}

\usepackage{amsmath}
\usepackage{bm}
\usepackage{threeparttable}
\usepackage{algorithm}
\usepackage{algpseudocode}
\usepackage{algorithmicx}

\usepackage{hyperref}       % hyperlinks
\usepackage{url}            % simple URL typesetting

\usepackage{natbib}

\newcommand{\bx}{\bm{x}}
\newcommand{\ba}{\bm{a}}
\newcommand{\bu}{\bm{u}}
\newcommand{\bB}{\bm{B}}
\newcommand{\bD}{\bm{D}}
\newcommand{\br}{\bm{r}}
\newcommand{\bTheta}{\bm{\Theta}}

\begin{document}

\title{Accelerating Hierarchical Sparse Predictive Coding with Hybrid Amortized Inference}
\author[1]{Kazuhisa Fujita}
\affil[1]{Department of Clinical Engineering, Komatsu University, 10-10 Doihara-Machi, Komatsu, Ishikawa, Japan 923-0921}
\corrauthor[1]{Kazuhisa Fujita}{kazu@spikingneuron.net}

%\author{Kazuhisa Fujita\\
%Komatsu University, 10-10 Doihara-Machi, Komatsu, Ishikawa, Japan 923-0921\\
%\texttt{kazu@spikingneuron.net}
%}

\begin{abstract}
Hierarchical predictive coding provides an interpretable framework for perception as error-driven inference in multi-layer models, while sparse coding imposes parsimonious latent representations through explicit sparsity constraints. Their combination yields hierarchical sparse predictive coding models with appealing computational and neuroscientific properties, but practical use is often limited by the cost of iterative latent inference. In such models, each input may require many recurrent refinement steps before a useful sparse representation is obtained, and this burden becomes more severe as the hierarchy deepens. We study this bottleneck by comparing training-and-inference procedures that share the same hierarchical sparse objective formulation and architecture but use different latent-inference mechanisms. The comparison includes classical iterative inference based on ISTA, an accelerated MFISTA reference, structurally informed amortized inference using a LISTA-style bottom-up encoder adapted to the hierarchical model, and a Hybrid procedure in which this fast amortized initialization is followed by a small number of corrective energy-based refinement steps. Each procedure is trained separately, allowing its inference mechanism to interact with dictionary learning and, where applicable, encoder learning. We measure the resulting reconstruction quality, sparsity, latency, and run-to-run variability across random seeds on static image benchmarks. The results show that Hybrid improves over pure amortization in the tested settings while remaining substantially faster than procedures based on long iterative inference.
\end{abstract}

\flushbottom
\maketitle
\thispagestyle{empty}

\section{Introduction}

Perception can be formulated as inference over latent causes. A useful model should recover those causes efficiently and reliably, while keeping the representation interpretable. Predictive coding frames this process as hierarchical error minimization: top-down predictions are compared with bottom-up inputs, and residual prediction errors propagate through the hierarchy \citep{Rao:1999,Friston:2009}. Sparse coding instead explains sensory signals with a small number of active coefficients, often using an overcomplete dictionary in classical formulations. This yields efficient and biologically meaningful representations; on natural images, sparse coding recovers localized, oriented, band-pass features resembling simple-cell receptive fields in primary visual cortex \citep{Olshausen:1996}.

These traditions are compatible. Predictive coding provides hierarchical generative structure and local error-driven inference; sparse coding provides an explicit sparsity prior and competition among units. Their combination has led to hierarchical sparse predictive coding models, most notably Sparse Deep Predictive Coding (SDPC), which integrates recurrent sparse inference within layers with feedforward and feedback interactions across layers \citep{Boutin:2021}. SDPC can learn oriented low-level receptive fields, more complex higher-level features, and context-sensitive effects such as contour integration, supporting hierarchical sparse predictive coding as both a computational and neuroscientific framework \citep{Boutin:2021}.

The main practical weakness is the inference engine. Standard formulations require $\ell_1$-regularized inverse problems to be solved for each input, often by ISTA (Iterative Shrinkage-Thresholding Algorithm) or related proximal methods \citep{Daubechies:2004,Beck:2009a,Beck:2009b}. These methods are principled, but they can be expensive: each new input may need many refinement steps before reaching a useful latent state. The cost becomes more serious in deeper hierarchies, where latency, instability, and repeated inner-loop optimization can dominate the system.

A natural improvement to this bottleneck is amortization, especially when the amortized map retains structure from the original sparse optimization problem. In sparse inference, Learning ISTA (LISTA) replaces many ISTA-like optimization steps with a learned feedforward network obtained by algorithm unrolling, yielding a structurally informed amortized predictor of approximate sparse codes \citep{Gregor:2010}. In predictive coding more broadly, hybrid predictive coding has shown how amortized and iterative inference can optimize one objective, combining fast feedforward initialization with recurrent refinement \citep{Tschantz:2023}.

We study how to equip hierarchical sparse predictive coding for static images with faster and more stable inference. We compare separately trained systems that use the same image architecture, sparse objective formulation, and evaluation protocol but differ in how latent inference is performed.

The paper contributes a shared-objective benchmark for hierarchical sparse predictive coding on static images, a unified comparison of procedures based on iterative, LISTA-style amortized, and Hybrid inference, and an empirical analysis of their quality, latency, and run-to-run variability.

\section{Related Work}

\paragraph{Sparse coding and dictionary learning}

Sparse coding provides a foundational framework for efficient representation learning by reconstructing inputs from a small number of active features. Classical formulations often use large, overcomplete dictionaries. The seminal work of \citet{Olshausen:1996} showed that sparse coding trained on natural images learns basis functions resembling receptive fields in primary visual cortex. It has therefore become established as both a useful feature-extraction method and a biologically plausible model of early vision. In this paper, however, sparse coding denotes the explicit sparsity constraint on latent coefficients; the compact layers used in our experiments are not necessarily overcomplete.

A large body of later work focused on scalable optimization and dictionary learning. Iterative shrinkage-thresholding methods and their accelerated variants became standard tools for solving the underlying $\ell_1$-regularized inverse problems \citep{Beck:2009a,Beck:2009b}. Related work developed online dictionary learning for sparse coding practical at larger scale \citep{Mairal:2010}. Other work also extended patch-based sparse coding to convolutional settings, which are better suited to images because they model translation structure more directly \citep{Bristow:2013}. These developments made sparse coding much more practical, but they did not remove its central inference bottleneck: obtaining sparse latent codes still usually requires iterative optimization for each input.

\paragraph{Predictive coding and free-energy-based inference}

Predictive coding is often formulated as a framework for approximate inference in hierarchical models \citep{Rao:1999,Friston:2005,Friston:2009}. In classical hierarchical formulations, higher cortical areas send predictions to lower areas, while lower areas return the mismatch between predicted and observed activity, that is, prediction errors \citep{Rao:1999,Friston:2005,Bastos:2012}. Recurrent exchanges between predictions and errors then iteratively update latent representations so that they better explain sensory input \citep{Rao:1999,Friston:2005}. This line of work was later generalized within the free-energy framework, in which perceptual inference is written as the minimization of a variational objective under a hierarchical model \citep{Friston:2009}. These formulations are attractive in computational neuroscience because both inference and learning can be described in terms of local message passing based on prediction errors \citep{Friston:2005,Bastos:2012}. Later work also showed that predictive-coding-style networks can approximate backpropagation under suitable conditions \citep{Whittington:2017,Rosenbaum:2022}. Related top-down modulation mechanisms have also been studied in biologically motivated visual models, including dynamic gating of task-relevant information in primary visual cortex and categorization-oriented interactions between higher and lower visual areas \citep{Kamiyama:2016,Abe:2018,Kashimori:2007}.

Here, predictive coding matters as both a neuroscientific theory and an algorithmic template. It provides hierarchical structure, local error-driven updates, and a natural iterative inference procedure. However, standard predictive coding networks require multiple recurrent steps per input, making deep inference costly and sometimes difficult to stabilize \citep{Tschantz:2023}.

\paragraph{Bridging predictive coding and sparse hierarchical models}

Several works have highlighted the compatibility between predictive coding and sparse coding. Predictive coding contributes hierarchical generative structure and error-driven inference; sparse coding contributes an explicit sparsity prior and competition among latent units. This has motivated models that embed sparse latent inference in predictive-coding-style top-down and bottom-up interactions.

A representative example is Sparse Deep Predictive Coding (SDPC), which combines predictive coding across layers with sparse recurrent inference within each layer in a hierarchical convolutional architecture \citep{Boutin:2021}. SDPC can learn oriented low-level receptive fields, structured higher-level representations, and context-sensitive effects such as contour integration. It therefore shows that predictive coding and sparse coding can be unified in a single model family for realistic image data. At the same time, it illustrates the main practical limitation of this approach: inference remains recurrent and expensive, especially as model depth and complexity increase \citep{Boutin:2021}.

\paragraph{Algorithm unrolling and learned sparse inference}

A separate but closely related line of work comes from algorithm unrolling for sparse inference. LISTA showed that the iterations of ISTA can be unfolded into a finite-depth neural network whose parameters are learned from data, thereby replacing slow iterative optimization with fast feedforward (i.e., amortized) inference that approximates sparse codes \citep{Gregor:2010}. Later work clarified and expanded this design space. For example, learned-step-size variants of unfolded ISTA were shown to be competitive with state-of-the-art learned sparse-inference networks when the target solutions are sufficiently sparse \citep{Ablin:2019}. Analytic LISTA (ALISTA) further showed that strong performance can be obtained even when only the step-sizes and thresholds are learned \citep{Liu:2019}.

Despite this progress, most unrolled sparse coding work has focused on single-layer sparse recovery rather than hierarchical predictive coding with explicit top-down generative interactions. This leaves an open question: how should fast amortized inference be combined with recurrent error-correcting refinement in a multi-layer sparse model?

\paragraph{Hybrid predictive coding and amortized-plus-iterative inference}

Recent work in predictive coding has begun to address this question from the viewpoint of amortized inference. Hybrid Predictive Coding (HPC) interprets feedforward inference as amortized inference and recurrent predictive-coding updates as iterative refinement under the same objective \citep{Tschantz:2023}. It formalizes a dual-process view: fast feedforward pathways provide cheap initial beliefs, while recurrent dynamics improve accuracy, context sensitivity, and robustness when more refinement is needed.

HPC provides a strong conceptual precedent, but it does not focus on hierarchical sparse coding with explicit $\ell_1$-type latent sparsity for static image representation. Furthermore, while HPC typically relies on unstructured black-box neural networks for the amortized pathway, our approach employs a structurally informed LISTA-style encoder with dictionary-derived initialization and learned shrinkage blocks. HPC establishes the general principle of combining amortized and iterative inference in predictive coding, while sparse coding provides tools for sparse objectives. This combination has been less directly studied for hierarchical sparse image models.

\paragraph{Positioning of the present work}

Our work is closest to SDPC on the modeling side and to LISTA and HPC on the inference side. The emphasis is different. Relative to SDPC, we vary the training-time and test-time inference engine while keeping the sparse model formulation and architecture fixed. Relative to LISTA and related unrolling methods, we study hierarchical latent inference rather than single-layer sparse recovery. Relative to HPC, we specialize amortized-plus-iterative inference to hierarchical sparse coding for static images, where reconstruction quality, sparsity, latency, and run-to-run variability can be compared under one sparse energy formulation. We therefore ask how amortized and iterative inference can be combined in this setting.

\section{Methods}

% hyperparameters:
% \beta_\ell: inter-layer coupling strength

\subsection{Overview}

We compare four training-and-inference procedures for the hierarchical sparse model defined below: (i) ISTA-style iterative inference as the basic iterative baseline \citep{Daubechies:2004,Beck:2009a}, (ii) MFISTA as a monotone accelerated iterative baseline \citep{Beck:2009a}, (iii) a shared-parameter LISTA-style amortized bottom-up encoder adapted to the hierarchical model \citep{Gregor:2010}, and (iv) a proposed Hybrid method in which this encoder provides an initialization that is further refined by a small number of ISTA-style steps. All procedures use the same hierarchical sparse objective formulation, architecture, sparsity coefficients, and coupling coefficients. However, they are trained independently and therefore generally produce different learned dictionaries and, where applicable, different encoder parameters.

\subsection{Hierarchical sparse model}

% n_0: number of pixels
% n_\ell: number of latent units in layer \ell

Let $\bx \in \mathbb{R}^{n_0}$ denote an observed image represented as a column vector, and let $\ba_\ell \in \mathbb{R}^{n_\ell}$ denote the latent code at layer $\ell \in \{1,\dots,L\}$. For each layer, $\bD_\ell \in \mathbb{R}^{n_{\ell-1} \times n_\ell}$ denotes a dictionary matrix. The hierarchical model is
\begin{align}
\bx &\approx \bD_1 \ba_1, \\
\ba_{\ell-1} &\approx \bD_\ell \ba_\ell, \qquad \ell = 2,\dots,L.
\end{align}
Equivalently, the layerwise residuals are
\begin{align}
\br_1 &= \bx - \bD_1 \ba_1, \\
\br_\ell &= \ba_{\ell-1} - \bD_\ell \ba_\ell, \qquad \ell = 2,\dots,L.
\end{align}
Thus, the first layer explains the sensory input, while deeper layers explain the activities of lower layers.

\subsection{Hierarchical sparse energy}

We define the single-sample energy as
\begin{equation}
E(\bx,\{\ba_\ell\},\{\bD_\ell\})
=
\frac12 \|\bx - \bD_1 \ba_1\|_2^2
+
\sum_{\ell=2}^{L} \frac{\beta_{\ell-1}}{2}\|\ba_{\ell-1} - \bD_\ell \ba_\ell\|_2^2
+
\sum_{\ell=1}^{L} \lambda_\ell \|\ba_\ell\|_1,
\label{eq:hsc_energy_single}
\end{equation}
where $\lambda_\ell > 0$ controls sparsity and $\beta_{\ell-1} > 0$ controls inter-layer coupling. The quadratic terms penalize input reconstruction error and inconsistency between adjacent latent layers. The $\ell_1$ terms encourage sparse latent codes like lasso regression. Thus, the objective extends standard sparse coding, reconstruction error plus sparsity, to a coupled hierarchical setting \citep{Olshausen:1996,Aberdam:2019}.

For a mini-batch $\{\bx^{(b)}\}_{b=1}^{B}$ where $B$ is the batch size and $\bx^{(b)}$ denotes the $b$-th sample in the batch, the training objective is the batch average of the same energy:
\begin{equation}
\mathcal{L}
=
\frac{1}{B}\sum_{b=1}^{B}
\left[
\frac12 \|\bx^{(b)} - \bD_1 \ba_1^{(b)}\|_2^2
+
\sum_{\ell=2}^{L} \frac{\beta_{\ell-1}}{2}\|\ba_{\ell-1}^{(b)} - \bD_\ell \ba_\ell^{(b)}\|_2^2
+
\sum_{\ell=1}^{L} \lambda_\ell \|\ba_\ell^{(b)}\|_1
\right].
\label{eq:hsc_batch_loss}
\end{equation}

\subsection{ISTA-style latent inference}

With dictionaries fixed, latent inference is posed as the minimization of Eq.~\eqref{eq:hsc_energy_single} with respect to $\{\ba_\ell\}_{\ell=1}^{L}$. Our basic iterative baseline is an ISTA-style proximal-gradient method: each refinement step computes layerwise gradients for the smooth part of the hierarchical energy and then applies soft-thresholding to the latent codes \citep{Daubechies:2004,Beck:2009a}.

Let $f$ denote the smooth part of Eq.~\eqref{eq:hsc_energy_single}, i.e.\ the same objective without the $\ell_1$ penalties. In column-vector form, the layerwise gradients are as follows.

For $\ell = 1$,
\begin{equation}
\label{eq:grad_a1}
\nabla_{\ba_1} f
=
\bD_1^\top(\bD_1 \ba_1 - \bx)
+
\beta_1(\ba_1 - \bD_2 \ba_2),
\end{equation}
when $L>1$, and the second term is omitted when $L=1$.

For $1 < \ell < L$,
\begin{equation}
\label{eq:grad_aell}
\nabla_{\ba_\ell} f
=
\beta_{\ell-1}\bD_\ell^\top(\bD_\ell \ba_\ell - \ba_{\ell-1})
+
\beta_\ell(\ba_\ell - \bD_{\ell+1} \ba_{\ell+1}).
\end{equation}

For $\ell=L$,
\begin{equation}
\label{eq:grad_aL}
\nabla_{\ba_L} f
=
\beta_{L-1}\bD_L^\top(\bD_L \ba_L - \ba_{L-1}).
\end{equation}

We then apply the proximal update
\begin{equation}
\ba_\ell \leftarrow \mathcal{S}_{\theta_\ell }\!\left(\ba_\ell - \eta_\ell \nabla_{\ba_\ell} f\right),
\label{eq:ista_update}
\end{equation}
where $\mathcal{S}_{\theta}$ is the soft-thresholding operator
\begin{equation}
\mathcal{S}_{\theta}(v) = \mathrm{sign}(v)\max(|v|-\theta,0),
\end{equation}
applied elementwise, and $\eta_\ell$ and $\theta_\ell$ are the step size and threshold for layer $\ell$, respectively.

In implementation, all layer gradients are evaluated at the current iterate, and the proximal updates are then applied within the same refinement step. The iterative baseline is therefore a block-Jacobi proximal-gradient refinement on the joint latent variables \citep{Beck:2009a,Beck:2009b,Parikh:2014,Xu:2018}, not a Gauss--Seidel or cyclic block-coordinate method \citep{Beck:2013,Xu:2018}. We use ``ISTA-style'' to emphasize the gradient-plus-soft-thresholding update form, while distinguishing this simultaneous multi-layer update from a literal cyclic application of classical single-block ISTA.

In standard proximal-gradient methods, fixed step sizes are determined by the Lipschitz constant of the full gradient \citep{Beck:2009a,Beck:2009b,Parikh:2014}. Similarly, block-coordinate variants use block-specific Lipschitz constants for each partial gradient \citep{Nesterov:2012,Richtarik:2014}. Following this principle, we determine layerwise step sizes based on the block-Lipschitz constants of the partial gradients for the smooth term.

For our smooth objective, the relevant block Hessians are
\begin{align}
\nabla^2_{a_1a_1} f &= \bD_1^\top \bD_1 + \beta_1 \bm{I}_{n_1} \qquad (L>1),\\
\nabla^2_{a_\ell a_\ell} f &= \beta_{\ell-1}\bD_\ell^\top \bD_\ell + \beta_\ell \bm{I}_{n_\ell}, \qquad (1<\ell<L),\\
\nabla^2_{a_La_L} f &= \beta_{L-1}\bD_L^\top \bD_L.
\end{align}
Taking the spectral norm of each block Hessian, the corresponding block-Lipschitz constants are
\begin{align}
L_1 &= \|\bD_1\|_\mathrm{spec}^2 + \beta_1 \qquad (L>1), \\
L_\ell &= \beta_{\ell-1}\|\bD_\ell\|_\mathrm{spec}^2 + \beta_\ell \qquad (1 < \ell < L), \\
L_L &= \beta_{L-1}\|\bD_L\|_\mathrm{spec}^2,
\end{align}
and for the single-layer case $L_1=\|\bD_1 \|_\mathrm{spec}^2$. Here, $\| \cdot \|_\mathrm{spec}$ is the spectral norm, so $\|\bD_\ell \|_\mathrm{spec}^2=\lambda_{\max}(\bD_\ell^\top\bD_\ell)$ is the largest eigenvalue of $\bD_\ell^\top\bD_\ell$. In practice, $\| \bD_\ell \|_\mathrm{spec}^2$ is estimated by 10 power iterations on $\bD_\ell^\top \bD_\ell$ from a fixed all-ones initial vector, followed by a Rayleigh-quotient estimate \citep{Demmel:1997,Golub:2013}.

%Rayleigh quotient: $R(M, x) = \frac{x^T M x}{x^T x}$

The actual step sizes are
\begin{equation}
\eta_\ell = \frac{\eta_{\mathrm{scale}}}{L_\ell},
\qquad
\theta_\ell = \eta_\ell \lambda_\ell,
\label{eq:stepsize_rule}
\end{equation}
where $\eta_{\mathrm{scale}}$ is a single global scaling hyperparameter that controls the overall speed of inference.
Because these layerwise updates are applied simultaneously to coupled latent variables, this blockwise rule should be read as a practical scaling choice rather than a monotonicity guarantee for the full joint objective.

In training, the step sizes are recomputed from the current detached dictionaries. When dictionaries are fixed at evaluation time, the same rule may be precomputed once and then reused for all inputs.

\subsection{LISTA-style amortized encoder}

For fast approximate inference, we use a bottom-up amortized encoder inspired by LISTA \citep{Gregor:2010}. Classical LISTA unfolds ISTA into a learned finite-depth sparse-coding network for a single sparse-coding problem. Here, we use the same idea as a structurally informed feed-forward inference module adapted to the hierarchical sparse model. This module is used both as the amortized initializer for Hybrid and as the LISTA-style inference method for hierarchical sparse coding in the comparisons below.

The encoder is composed of $L$ layerwise blocks. Each block maps the current layer input to an approximate latent code for that layer, and the resulting code is then passed to the next block in a bottom-up manner:
\begin{align}
\ba_1 &= f_1(\bx), \\
\ba_2 &= f_2(\ba_1), \\
&\vdots \\
\ba_L &= f_L(\ba_{L-1}).
\end{align}
Thus, the encoder defines a feed-forward LISTA-style amortized approximation to the hierarchical latent codes. It should not be read as an exact unrolling of proximal-gradient inference for the full coupled hierarchical objective. Instead, each block produces a bottom-up sparse-code estimate using an ISTA-like shrinkage recurrence. The reconstruction error and inter-layer consistency errors are not used explicitly inside this feed-forward encoder; they enter explicitly during Hybrid refinement and, in LISTA and Hybrid, through the sparse energy used to train the encoder parameters.
%LISTAではフィードフォワードで処理するから、誤差を計算しない。

In the main experiments, we use the \emph{shared-parameter} variant. For a given layer input $\bu_\ell$, one encoder block first computes
\begin{equation}
\bB_\ell = W_{x,\ell} \bu_\ell,
\end{equation}
where $W_{x,\ell}$ is an encoder weight matrix, and then applies a fixed number of shrinkage stages:
\begin{align}
\ba^{(0)}_{\ell} &= \mathcal{S}_{\bm{\vartheta}_\ell}(\bB_\ell), \\
\ba^{(t+1)}_{\ell} &= \mathcal{S}_{\bm{\vartheta}_\ell}\!\left(\bB_\ell + W_{a,\ell} \ba^{(t)}_{\ell}\right),
\qquad t=0,\dots,K-2,
\label{eq:shared_lista_block}
\end{align}
where $K$ is the total number of LISTA-style shrinkage stages in the block, the final output is $\ba^{(K-1)}_{\ell}$, and $W_{a,\ell}$ is another encoder weight matrix. Thus, $K=1$ means that only the initial shrinkage stage is applied. Here, $\mathcal{S}_{\bm{\vartheta}_\ell}$ denotes elementwise soft-thresholding with a learned threshold vector $\bm{\vartheta}_\ell\in\mathbb{R}_{+}^{n_\ell}$, distinct from the scalar ISTA refinement threshold $\theta_\ell$ in Eq.~\eqref{eq:stepsize_rule}. Each entry of $\bm{\vartheta}_\ell$ is the threshold for one latent unit, so the block learns one threshold per latent unit. The same parameters $(W_{x,\ell}, W_{a,\ell}, \bm{\vartheta}_\ell)$ are shared across all internal shrinkage stages within a block. To ensure non-negative thresholds, we reparameterize
\begin{equation}
\bm{\vartheta}_\ell = \mathrm{softplus}(\bm{\rho}_\ell),
\end{equation}
where $\bm{\rho}_\ell\in\mathbb{R}^{n_\ell}$ is a learnable unconstrained encoder parameter and softplus is applied elementwise.

During training and evaluation, LISTA-style amortized inference uses the current encoder weights and thresholds directly. Its shrinkage stages in Eq.~\eqref{eq:shared_lista_block} do not explicitly compute or use an ISTA step size. Instead, the encoder weight matrices $W_{x,\ell}$ and $W_{a,\ell}$, together with the threshold vector $\bm{\vartheta}_\ell=\operatorname{softplus}(\bm{\rho}_\ell)$, are learned directly from the training objective. Pure LISTA returns this feed-forward code directly. In the Hybrid variant, the same forward pass produces an initial latent code, then refines it with ISTA-style updates under the full hierarchical energy.

\subsection{Hybrid inference}

The proposed method combines amortized latent-code initialization with iterative refinement. Here ``initialization'' means the starting latent codes for the current input, not initialization of encoder parameters. Hybrid first runs the LISTA-style amortized encoder and then applies ISTA-style refinement. Given $\bx$, the bottom-up initialization is
\begin{equation}
\{\ba_\ell^{(0)}\}_{\ell=1}^{L} = \mathrm{LISTA}(\bx;\phi),
\end{equation}
where $\phi$ denotes the encoder parameters. This forward pass gives a fast approximate latent code. Other amortized architectures could in principle be substituted.

Starting from this initialization, we then perform $T_{\mathrm{ref}}$ ISTA-style refinement steps using Eq.~\eqref{eq:ista_update}. The final inferred codes are
\begin{equation}
\hat{\ba}_\ell = \ba_\ell^{(T_{\mathrm{ref}})}, \qquad \ell=1,\dots,L.
\end{equation}
These refinement steps apply proximal updates derived from the full hierarchical sparse energy to the initial amortized codes.

The amortized encoder provides a low-cost initial estimate. LISTA returns this estimate directly, whereas Hybrid follows it with a small number of objective-based recurrent updates.

\subsection{Training procedure and gradient flow}

\paragraph{Alternating training view}
Training alternates between latent inference and parameter updates. This design is best viewed as an alternating-optimization-style approximation:
\begin{enumerate}
\item infer latent codes while holding dictionaries fixed;
\item evaluate the batch loss using the learnable dictionaries;
\item update dictionaries, and when applicable encoder parameters, using that final loss.
\end{enumerate}
The inferred codes are therefore treated as the result of optimizing latent variables with fixed dictionaries. Dictionary updates use backpropagation from the final sparse-coding loss, but the dictionaries are detached during latent refinement, so dictionary gradients are not propagated through the refinement steps. This reduces memory use and improves training stability.

\paragraph{Dictionary gradient flow}
During latent inference, the dictionaries are treated as fixed. In the implementation, the learnable dictionaries $\{\bD_\ell\}_{\ell=1}^{L}$ are first converted into detached inference dictionaries,
\begin{equation}
\tilde{\bD}_\ell = \operatorname{stopgrad}(\bD_\ell), \qquad \ell=1,\dots,L,
\end{equation}
and the inference engine receives $\{\tilde{\bD}_\ell\}_{\ell=1}^{L}$ rather than the original parameters. Conceptually, the inferred codes are computed as
\begin{equation}
\{\hat{\ba}_\ell\}_{\ell=1}^{L}
=
\operatorname{Infer}
\left(
\bx;
\{\tilde{\bD}_\ell\}_{\ell=1}^{L}
\right).
\end{equation}
Thus, inference uses the current dictionary values, but the computation graph does not keep a gradient path from the inferred codes back to the dictionary parameters through the inference trajectory.

After inference, the final mini-batch loss is evaluated with the original learnable dictionaries, not with the detached copies:
\begin{equation}
\mathcal{L}
=
E
\left(
\bx,
\{\hat{\ba}_\ell\}_{\ell=1}^{L},
\{\bD_\ell\}_{\ell=1}^{L}
\right).
\end{equation}
Dictionary learning is therefore driven by the explicit dependence of the final loss on $\{\bD_\ell\}_{\ell=1}^{L}$. If the detached dictionaries were also used in this final loss, the loss would contain no dictionary-gradient path and the dictionaries would not be learned from the reconstruction and inter-layer consistency terms. The inferred codes are treated as constants when updating dictionaries. Equivalently, the update differentiates the final energy with respect to dictionary parameters, but not through the iterative inference procedure.

\paragraph{Encoder gradient flow}
In Hybrid, the LISTA-style encoder provides the initial codes for refinement. Because these initial codes are not detached, gradients flow from the final loss through the refined codes and the refinement iterations back to the encoder parameters. The dictionaries used within refinement are detached, preventing gradients from flowing through the refinement trajectory to the dictionary parameters. Instead, dictionary gradients are obtained by evaluating the final loss with the refined codes and the original learnable dictionaries. Thus, the encoder is trained to produce initial codes that lead, after refinement, to low-loss solutions. The appendix-only Hybrid-MFISTA ablation follows the same gradient convention.

\paragraph{Learned parameter groups}
The learnable dictionary parameters are
\begin{equation}
\bTheta_D = \{\bD_\ell\}_{\ell=1}^{L}.
\end{equation}
For LISTA and Hybrid, the model also contains amortized encoder parameters, denoted by
\begin{equation}
\bTheta_E = \phi .
\end{equation}
ISTA and MFISTA do not contain such encoder parameters, since they compute latent codes only by iterative inference.

Parameter learning uses standard gradient-based optimization. In all modes, $\bTheta_D$ is updated from the final loss. In LISTA and Hybrid, $\bTheta_E$ is updated from the same loss. Both parameter groups are optimized with Adam using separate learning rates. Thus, the same hierarchical sparse objective trains the dictionaries and, when present, the amortized encoder.

LISTA and Hybrid are trained as separate models. They share the same encoder architecture and initialization rule, but their learned encoder parameters are not shared. The appendix-only Hybrid-MFISTA ablation is also trained separately, using the Hybrid encoder setup while replacing only the refinement rule.

\paragraph{Dictionary normalization}
After each parameter update, all dictionary atoms are normalized:
\begin{equation}
\bm{d}_{\ell,m}
\leftarrow
\frac{\bm{d}_{\ell,m}}{\max(\|\bm{d}_{\ell,m}\|_2,\delta)},
\end{equation}
where $\bm{d}_{\ell,m}$ denotes the $m$-th atom of dictionary $\bD_\ell$ and $\delta>0$ is a small constant.

\paragraph{Encoder initialization}
Before training, the generative dictionaries are randomly initialized and column-normalized. For LISTA and Hybrid, the LISTA-style encoder parameters are then initialized once from these dictionaries. This dictionary-derived initialization is not repeated during later inference or evaluation. For layer $\ell$, let $\bD_\ell$ be the initial local dictionary and let
\begin{equation}
L_\ell^{\mathrm{enc}} \approx \|\bD_\ell\|_\mathrm{spec}^2
\end{equation}
be the spectral estimate used for this encoder initialization. This estimate uses the same 10-step power-iteration procedure as the inference step-size rule. We set
\begin{equation}
\eta_\ell^{\mathrm{enc}} = \frac{\eta_{\mathrm{scale}}}{L_\ell^{\mathrm{enc}}},
\end{equation}
and initialize
\begin{align}
W_{x,\ell} &\leftarrow \eta_\ell^{\mathrm{enc}} \bD_\ell^\top, \\
W_{a,\ell} &\leftarrow \bm{I}_{n_\ell} - \eta_\ell^{\mathrm{enc}} \bD_\ell^\top \bD_\ell, \\
\bm{\vartheta}_\ell &\leftarrow \eta_\ell^{\mathrm{enc}} \lambda_\ell \bm{1}_{n_\ell}.
\end{align}
In the softplus parameterization, this last assignment is implemented by setting every entry of $\bm{\rho}_\ell$ so that $\mathrm{softplus}(\rho_{\ell,j})=\eta_\ell^{\mathrm{enc}}\lambda_\ell$. Thus, all latent-unit thresholds have the same initial value, but they are separate parameters and need not remain equal after training.

After this one-time initialization, the encoder is not reinitialized as the dictionaries change. Instead, $W_{x,\ell}$, $W_{a,\ell}$, and $\bm{\rho}_\ell$ are learned by directly minimizing the hierarchical sparse loss in Eq.~\eqref{eq:hsc_batch_loss}, rather than by reproducing latent codes obtained from an iterative solver. Thus, $L_\ell^{\mathrm{enc}}$ and $\eta_\ell^{\mathrm{enc}}$ serve only as initialization quantities, not as step-size parameters used during later LISTA inference.

\paragraph{Overall training procedure}
Training proceeds as follows:
\begin{enumerate}
    \item sample a mini-batch of images;
    \item flatten images into vectors and optionally remove the per-sample DC component;
    \item infer latent codes using one of the four inference engines: ISTA-style iterative inference, MFISTA, shared LISTA, or Hybrid;
    \item compute the batch-averaged hierarchical sparse loss in Eq.~\eqref{eq:hsc_batch_loss};
    \item update dictionaries and, when applicable, encoder parameters by gradient-based optimization;
    \item normalize all dictionary atoms.
\end{enumerate}

The architecture, dictionary capacity, sparsity coefficients, and evaluation protocol are fixed across methods. The procedures differ in the inference engine used in Step 3 and consequently in their learned dictionaries; LISTA and Hybrid additionally introduce amortized encoder parameters trained under the same sparse objective.

\subsection{Algorithm summaries}

The Hybrid inference and training procedures are summarized in Algorithms~\ref{alg:hybrid_inference} and~\ref{alg:training_procedure}, respectively. The gradients above use column-vector notation for a single sample, whereas the algorithms follow the implementation convention of storing batch samples as rows.

% n_0: input dimension
% n_\ell: latent dimension at layer \ell
% B: batch size
\begin{algorithm}[htbp]
\small
\caption{Hybrid inference for $L$-layer hierarchical sparse coding}
\label{alg:hybrid_inference}
\begin{algorithmic}[1]
\Require input batch $\bm{X} \in \mathbb{R}^{B \times n_0}$, dictionaries $\{\bD_\ell\}_{\ell=1}^{L}$ with $\bD_\ell \in \mathbb{R}^{n_{\ell-1}\times n_\ell}$, sparsity weights $\{\lambda_\ell\}_{\ell=1}^{L}$, coupling weights $\{\beta_\ell\}_{\ell=1}^{L-1}$, amortized encoder, LISTA-style shrinkage stages $K$, refinement steps $T_{\mathrm{ref}}$, scale parameter $\eta_{\mathrm{scale}}$
\Ensure latent codes $\{\ba_\ell\}_{\ell=1}^{L}$

\Statex \textbf{Step 1: amortized bottom-up initialization with shared LISTA blocks}
\State $\bm{U} \gets \bm{X}$
\For{$\ell = 1$ to $L$}
    \State $\bm{B}_\ell \gets \bm{U} W_{x,\ell}^{\top}$
    \State $\ba_\ell \gets \mathcal{S}_{\bm{\vartheta}_\ell}(\bm{B}_\ell)$
    \For{$k = 1$ to $K-1$}
        \State $\ba_\ell \gets \mathcal{S}_{\bm{\vartheta}_\ell}(\bm{B}_\ell + \ba_\ell W_{a,\ell}^{\top})$
    \EndFor
    \State $\bm{U} \gets \ba_\ell$
\EndFor

\Statex \textbf{Step 2: layerwise step sizes}
\For{$\ell = 1$ to $L$}
    \If{$\ell = 1$}
        \State $L_1 \gets \|\bD_1\|_\mathrm{spec}^2$ % power iteration estimate
        \If{$L > 1$}
            \State $L_1 \gets L_1 + \beta_1$
        \EndIf
    \ElsIf{$\ell = L$}
        \State $L_L \gets \beta_{L-1}\|\bD_L\|_\mathrm{spec}^2$
    \Else
        \State $L_\ell \gets \beta_{\ell-1}\|\bD_\ell\|_\mathrm{spec}^2 + \beta_\ell$
    \EndIf
    \State $\eta_\ell \gets \eta_{\mathrm{scale}}/L_\ell$
    \State $\theta_\ell \gets \eta_\ell \lambda_\ell$
\EndFor

\Statex \textbf{Step 3: ISTA-style refinement with simultaneous gradient evaluation}
\For{$t = 1$ to $T_{\mathrm{ref}}$}
    \For{$\ell = 1$ to $L$}
        \If{$\ell = 1$}
            \State $\bm{X}_{\mathrm{rec}} \gets \ba_1\bD_1^\top$
            \State $\bm{g}_\ell \gets (\bm{X}_{\mathrm{rec}} - \bm{X})\bD_1$
            \If{$L > 1$}
                \State $\bm{P}_{\mathrm{upper}} \gets \ba_2\bD_2^\top$
                \State $\bm{g}_\ell \gets \bm{g}_\ell + \beta_1(\ba_1 - \bm{P}_{\mathrm{upper}})$
            \EndIf
        \ElsIf{$\ell = L$}
            \State $\bm{P}_{\mathrm{lower}} \gets \ba_L\bD_L^\top$
            \State $\bm{g}_\ell \gets \beta_{L-1}(\bm{P}_{\mathrm{lower}} - \ba_{L-1})\bD_L$
        \Else
            \State $\bm{P}_{\mathrm{lower}} \gets \ba_\ell\bD_\ell^\top$
            \State $\bm{P}_{\mathrm{upper}} \gets \ba_{\ell+1}\bD_{\ell+1}^\top$
            \State $\bm{g}_{\mathrm{lower}} \gets \beta_{\ell-1}(\bm{P}_{\mathrm{lower}} - \ba_{\ell-1})\bD_\ell$
            \State $\bm{g}_{\mathrm{upper}} \gets \beta_\ell(\ba_\ell - \bm{P}_{\mathrm{upper}})$
            \State $\bm{g}_\ell \gets \bm{g}_{\mathrm{lower}} + \bm{g}_{\mathrm{upper}}$
        \EndIf
    \EndFor
    \For{$\ell = 1$ to $L$}
        \State $\ba_\ell \gets \mathcal{S}_{\theta_\ell }(\ba_\ell - \eta_\ell \bm{g}_\ell)$
    \EndFor
\EndFor

\State \Return $\{\ba_\ell\}_{\ell=1}^{L}$
\end{algorithmic}
\end{algorithm}

\begin{algorithm}[htbp]
\small
\caption{Training procedure used in the implementation}
\label{alg:training_procedure}
\begin{algorithmic}[1]
\Require training set $\mathcal{D}$, model mode $m$, dictionaries $\{\bD_\ell\}_{\ell=1}^{L}$, optional encoder parameters $\phi$
\Statex \hspace{\algorithmicindent}
$m \in \{\texttt{ista}, \texttt{mfista}, \texttt{lista}, \texttt{hybrid},$
\Statex \hspace{\algorithmicindent}
$\texttt{hybrid\_mfista}\}$
\Ensure trained dictionaries $\{\bD_\ell\}_{\ell=1}^{L}$ and, when applicable, encoder parameters $\phi$

\State initialize dictionaries $\{\bD_\ell\}$ randomly
\For{$\ell = 1$ to $L$}
    \State normalize columns of $\bD_\ell$
\EndFor

\If{$m$ uses a LISTA-style encoder}
    \State initialize encoder parameters from current dictionaries
\EndIf

\For{epoch = 1 to $N_{\mathrm{epoch}}$}
    \For{mini-batch $\bm{X}$ from $\mathcal{D}$}
        \State flatten input images into row vectors $\bm{X} \in \mathbb{R}^{B\times n_0}$
        \If{DC centering is enabled}
            \State subtract per-sample mean from each row of $\bm{X}$
        \EndIf

        \State $\tilde{\bD}_\ell \gets \mathrm{detach}(\bD_\ell)$ for all $\ell$
        \State infer latent codes $\{\ba_\ell\}$ using the selected inference engine with $\{\tilde{\bD}_\ell\}$

        \State compute batch-averaged energy with the original, non-detached dictionaries
        \Statex \hspace{\algorithmicindent}
        $\mathcal{L}
        =
        \frac{1}{B}\sum_{b=1}^{B}
        \left[
        \frac12 \|\bx^{(b)} - \bD_1 \ba_1^{(b)}\|_2^2
        +
        \sum_{\ell=2}^{L}\frac{\beta_{\ell-1}}{2}\|\ba_{\ell-1}^{(b)} - \bD_\ell \ba_\ell^{(b)}\|_2^2
        +
        \sum_{\ell=1}^{L}\lambda_\ell \|\ba_\ell^{(b)}\|_1
        \right]$

        \State take an Adam step on $\mathcal{L}$ to update $\{\bD_\ell\}_{\ell=1}^{L}$
        \If{encoder exists}
            \State take an Adam step on $\mathcal{L}$ to update $\phi$
        \EndIf

        \For{$\ell = 1$ to $L$}
            \State normalize columns of $\bD_\ell$
        \EndFor
    \EndFor
\EndFor

\State \Return $\{\bD_\ell\}_{\ell=1}^{L}$ and $\phi$
\end{algorithmic}
\end{algorithm}

\section{Experimental Setup}
\label{sec:experiments}

The experiments compare the learned systems produced by the four inference mechanisms. We specify the datasets, shared hyperparameters, inference-budget sweeps, latency measurements, and other plotted quantities used to generate the reported figures. The experimental design is organized around three questions:
\begin{enumerate}
    \item How do ISTA, MFISTA, LISTA, and the proposed Hybrid method compare in final test loss, reconstruction error, and inference latency?
    \item How does this comparison change as the hierarchical model becomes deeper?
    \item Are the observed trends stable under changes in sparsity strength and implementation-level choices in the refinement step size?
\end{enumerate}

\subsection{Datasets}

The main experiments use three static-image benchmarks: MNIST,
Fashion-MNIST, and CIFAR-10 Gray. MNIST provides a simple reference
regime, Fashion-MNIST adds more structured grayscale images, and
CIFAR-10 Gray, a grayscale version of CIFAR-10, tests whether the same
inference trade-off persists on compact natural images.
Appendix~\ref{app:natural_image_results} adds BSDS500 Patch as a
natural-image patch check connected to the classical sparse-coding
setting; it consists of $16 \times 16$ patches derived from the BSDS500
natural-image dataset \citep{Arbelaez:2011}.

For all datasets, 8-bit pixel intensities are converted to 32-bit
floating-point values and divided by 255, scaling the model inputs to
$[0,1]$ before flattening.
For CIFAR-10 Gray, RGB images are converted to grayscale using
$Y = 0.2989R + 0.5870G + 0.1140B$.

For MNIST, Fashion-MNIST, and CIFAR-10 Gray, we keep the official test
split untouched and reserve $10\%$ of the official training split for
validation using a fixed random seed of $0$. The validation split is used
for model monitoring and epoch-wise convergence curves. Final quality
metrics for method comparisons are computed on the held-out test split. The
dedicated latency-only pass also uses test batches. For the appendix-only
BSDS500 Patch experiment, we sample $16 \times 16$ patches from the
official train, validation, and test image splits, using 54000, 6000,
and 10000 patches, respectively.

Table~\ref{tab:dataset_summary} summarizes the datasets and sample counts
used in the experiments.

\begin{table}[t]
\centering
\caption{Dataset summary.}
\label{tab:dataset_summary}
\small
\begin{tabular}{lllll}
\hline
Dataset       & Resolution    & train size  & validation size & test size \\
\hline
MNIST         & $28 \times 28$ & 54000      & 6000 & 10000 \\
Fashion-MNIST & $28 \times 28$ & 54000      & 6000 & 10000 \\
CIFAR-10 Gray & $32 \times 32$ & 45000      & 5000 & 10000 \\
BSDS500 Patch & $16 \times 16$ & 54000 & 6000 & 10000 \\
\hline
\end{tabular}
\end{table}

\subsection{Compared methods and protocol}

The primary comparison includes ISTA, MFISTA, LISTA, and Hybrid. ISTA and MFISTA are iterative reference methods, LISTA denotes the LISTA-style amortized encoder baseline adapted to the hierarchical model, and Hybrid uses an amortized initialization followed by ISTA-style refinement. Hybrid-MFISTA, which replaces the Hybrid refinement stage with MFISTA-style refinement, is not part of the primary comparison; it is treated only as an appendix additional ablation and reported in Table~\ref{tab:hybrid_mfista_supp}. The algorithmic details of these methods are given in the Methods section.

ISTA and MFISTA serve as iterative references at the high-computation end of the quality--latency trade-off. Their default iteration counts define practical operating points rather than the best achievable performance, while separate budget sweeps evaluate the effect of additional computation. The main question is whether Hybrid improves on pure amortized inference with only a small amount of iterative refinement.

All four primary methods use the same sparse energy, data splits, dictionary architecture, optimizer settings, and evaluation code. The ablation experiments then vary one named factor at a time, such as hierarchy depth (Fig.~\ref{fig:depth_scaling_main}), sparsity strength (Fig.~\ref{fig:sparsity_sweep}), or Hybrid implementation choices (Fig.~\ref{fig:ablation}).

\subsection{Fixed hyperparameters and ablations}

The default model is a two-layer hierarchy with latent dimensions $(256,64)$. Thus the first dictionary maps the flattened input to 256 coefficients, and the second maps 256 coefficients to 64 coefficients; the input dimension is dataset dependent (784 for $28\times28$ grayscale images, 1024 for CIFAR-10 Gray, and 256 for $16\times16$ BSDS500 patches). Unless an ablation changes them, the sparsity weights are tied across layers, $\lambda_\ell=0.05$, and the inter-layer coupling is $\beta_1=1.0$. Dictionary parameters and, when present, encoder parameters are optimized with Adam using learning rates $10^{-3}$ for both $\bTheta_D$ and $\bTheta_E$. We use $\eta_{\mathrm{scale}}=1.0$, batch size 256, 25 training epochs, and no per-sample DC centering.

We selected these defaults as common settings that were stable across the tested datasets and methods. In particular, $\lambda=0.05$ represents a working trade-off between sparsity and reconstruction quality; the sparsity sweep examines how changing this value affects both quantities.

The depth-scaling experiment uses the following script-defined layer schedules:
\begin{center}
\small
\begin{tabular}{ll}
$L=1$ & $(256)$ \\
$L=2$ & $(256,64)$ \\
$L=4$ & $(256,192,128,64)$ \\
$L=6$ & $(256,224,192,160,128,64)$ \\
$L=8$ & $(256,240,224,192,160,128,96,64)$ \\
\end{tabular}
\end{center}
This sweep keeps $\lambda_\ell=0.05$ and $\beta_\ell=1.0$ wherever applicable. The sparsity ablation sweeps tied coefficients $\lambda \in \{0.02,0.05,0.10,0.20\}$, and the step-size ablation sweeps $\eta_{\mathrm{scale}} \in \{0.5,1.0,1.5\}$. Separate inference-budget sweeps for ISTA, MFISTA, LISTA, and Hybrid are used for the quality-versus-budget comparison in Fig.~\ref{fig:pareto_main} and the budget--latency mapping in Fig.~\ref{fig:budget_latency_main}, so the comparison is not tied only to the single default inference budget in Table~\ref{tab:default_hparams}. For ISTA and MFISTA, each additional inference step also adds computation time; the latency plot therefore shows which quality gains come at the cost of slower inference.

The \emph{inference budget} denotes the number of method-specific inference steps used to produce a latent code. For ISTA and MFISTA, it is the number of iterative update steps. For LISTA-style amortized inference, it is the number of learned shrinkage stages $K$ within each layerwise encoder block, rather than an unrolling depth of the full coupled hierarchical objective. For Hybrid, it is specified by the pair $(K,T_{\mathrm{ref}})$. In the main quality-versus-budget sweep, the Hybrid amortized depth is fixed at $K=1$, and only $T_{\mathrm{ref}}$ is varied. The separate Hybrid allocation analysis in Section~\ref{sec:hybrid_allocation} varies $K$ and $T_{\mathrm{ref}}$ independently.

For readability, the default operating point used throughout the main comparisons is summarized in Table~\ref{tab:default_hparams}.

\begin{table}[t]
\centering
\caption{Default experimental hyperparameters used in the main comparisons.}
\label{tab:default_hparams}
\small
\resizebox{0.8\linewidth}{!}{
\begin{tabular}{ll}
\hline
Component & Default setting \\
\hline
Hierarchy depth & $L=2$ \\
Layer dimensions & $(256, 64)$ \\
Sparsity weights & $\lambda_1=\lambda_2=0.05$ \\
Inter-layer coupling & $\beta_1=1.0$ \\
Dictionary optimizer & Adam, learning rate $10^{-3}$ \\
Encoder optimizer & Adam, learning rate $10^{-3}$ \\
Step-size scale & $\eta_{\mathrm{scale}}=1.0$ \\
Batch size & 256 \\
Training length & 25 epochs \\
DC centering & off by default \\
ISTA inference budget & 50 inference steps \\
MFISTA inference budget & 20 inference steps \\
LISTA inference budget & $K=1$ shrinkage stage \\
Hybrid inference budget & $(K, T_{\mathrm{ref}})=(1, 5)$ \\
Random seeds & target core sweeps: $\{0,1,2\}$; dedicated latency evaluation: $\{0\}$ \\
\hline
\end{tabular}
}
\end{table}

\subsection{Evaluation metrics}

Because all methods optimize or approximate the same sparse generative objective, the primary plotted quantities are two quality metrics and one timing metric. The sparsity sweep additionally reports the mean active fraction. For final method comparisons, the quality metrics are computed on the held-out test split after training. The timing metric is computed separately on test batches by the dedicated latency-only protocol.

The first quality metric, \emph{test loss}, is the full batch-averaged hierarchical sparse energy in Eq.~\eqref{eq:hsc_batch_loss}, including the input reconstruction term, the inter-layer consistency terms, and the $\ell_1$ sparsity penalties. The second, \emph{reconstruction error}, retains only the input-space term $\frac12\|\bx-\bD_1\ba_1\|_2^2$. Unless otherwise stated, final test loss and reconstruction error are aggregated over three runs with seeds $0$, $1$, and $2$ and plotted as mean values with standard deviation error bars. Here, run-to-run variability refers to empirical variation in these final metrics across the three independently trained seed runs; it is not a formal convergence guarantee or a measure of inference-dynamics stability for a fixed trained model. The convergence figures plot these two quantities on the validation split over epochs. In the sparsity sweep, we additionally plot the mean active fraction, defined as the unweighted mean across latent layers of the fraction of inferred coefficients whose magnitude exceeds $10^{-3}$.

\emph{Latency} denotes the measured wall-clock inference time in ms/sample. When latency is reported, the value is the median ms/sample from the dedicated latency evaluation matched to the same method and inference-budget setting. Latency was measured using freshly initialized, untrained model instances with the same architectures and inference budgets; the trained quality checkpoints were not loaded. This evaluation uses seed $0$, batch size 1, 100 warm-up batches, and at most 500 timed test batches. It is intended to compare the computational cost of the inference procedures, whereas quality metrics come from separate training-based runs.

\subsection{Implementation details}

All methods are implemented in Python using PyTorch. The core sparse-coding implementation is publicly available at \url{https://github.com/KazuhisaFujita/HSC}.

Only the dedicated latency evaluation is run under a fixed hardware and execution setting: a single-threaded run on a machine with 64 GB RAM and an AMD Ryzen 9900X CPU, using the same batch-size convention and summary format across methods. The training-based experiments used for final test quality metrics were run with parallel processing on different hardware environments. Therefore, the paper uses the dedicated latency pass for timing comparisons, while quality metrics are taken from the corresponding training-based runs.

\section{Results}
\label{sec:results}

\subsection{Quality versus inference budget and latency}

\begin{figure}[htbp]
\centering
\includegraphics[width=0.95\linewidth]{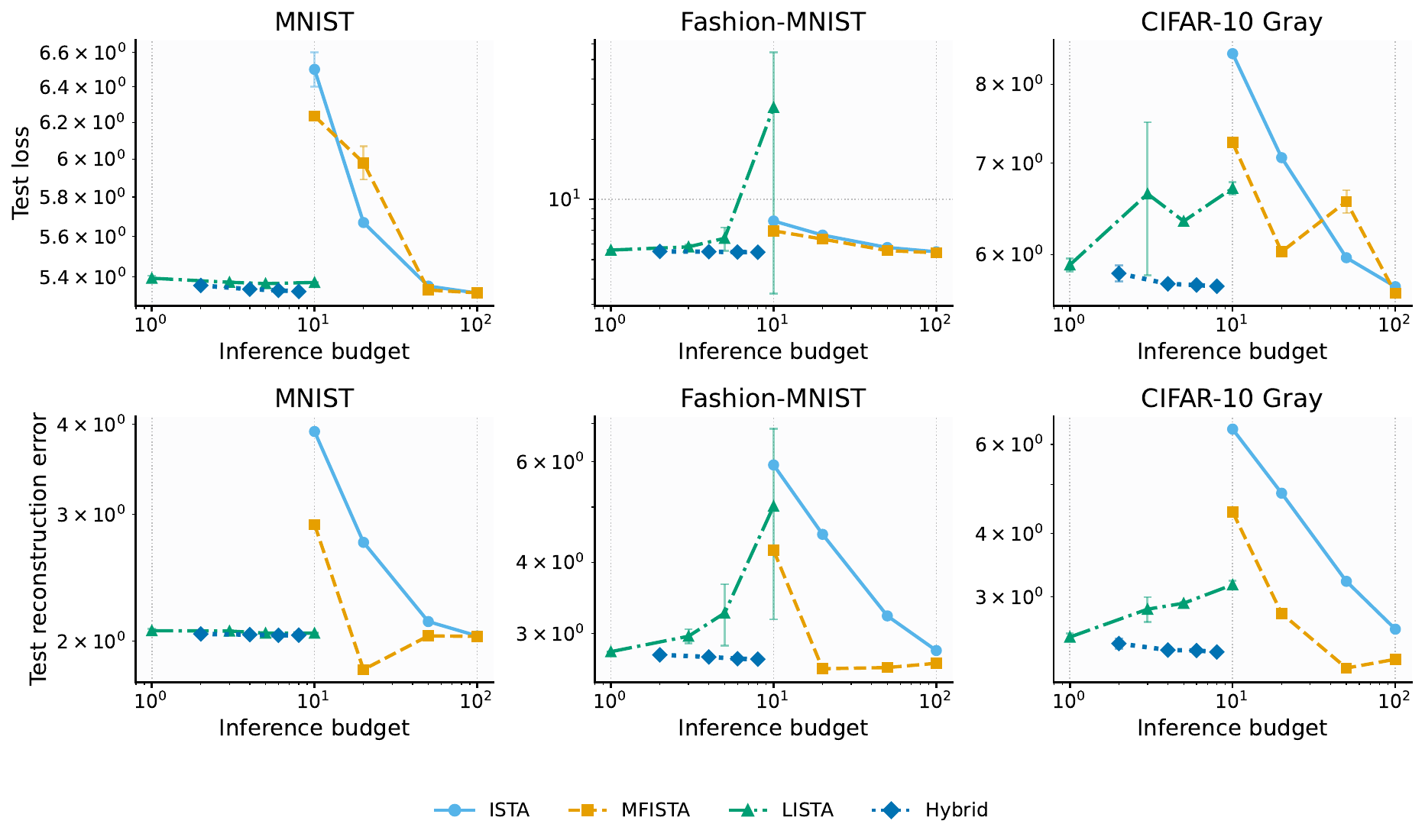}
\caption{Quality versus inference budget for ISTA, MFISTA, LISTA, and Hybrid. Columns show datasets; rows show test loss and reconstruction error. For Hybrid, the amortized depth is fixed at $K=1$ and the curve varies only $T_{\mathrm{ref}}$. Markers show means over three runs with seeds $0$, $1$, and $2$, and error bars indicate standard deviation across seeds.}
\label{fig:pareto_main}
\end{figure}

Figure~\ref{fig:pareto_main} compares the four procedures at a fixed two-layer hierarchy; every method--budget setting is trained separately. Columns correspond to datasets, and rows show the two test metrics: test loss and reconstruction error. The horizontal axis is the method-specific inference budget. For ISTA and MFISTA this is the number of iterative updates, for LISTA-style amortized inference it is the number of learned shrinkage stages per layerwise encoder block, and for the displayed Hybrid curve it is $K+T_{\mathrm{ref}}$ with the amortized depth fixed at $K=1$. Each marker is the mean over three runs at a given inference budget, with error bars showing between-seed variation in quality.

The main pattern is that LISTA-style amortized inference gives the lowest-budget solution, but additional layerwise shrinkage stages do not reliably close the quality gap. On Fashion-MNIST and CIFAR-10 Gray in particular, configurations with more amortized stages can have worse test loss or reconstruction error than the $K=1$ setting. By contrast, the $K=1$ Hybrid configurations outperform the corresponding amortized baseline with only a few corrective updates. Across all three datasets, these configurations have lower test loss and reconstruction error than $K=1$ LISTA, and the results are usually maintained or slightly improved as $T_{\mathrm{ref}}$ increases from 1 to 7.

\begin{figure}[htbp]
\centering
\includegraphics[width=0.95\linewidth]{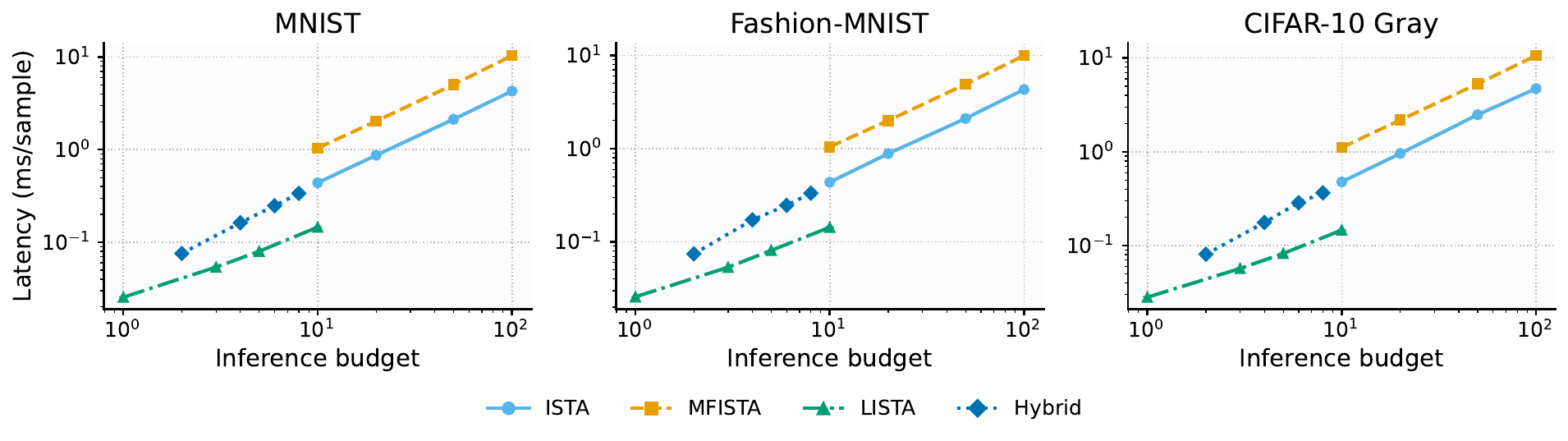}
\caption{Measured latency as a function of inference budget for the same fixed-depth settings used in Fig.~\ref{fig:pareto_main}. For Hybrid, the amortized depth is fixed at $K=1$ and the budget changes through $T_{\mathrm{ref}}$. Markers show median latency from the dedicated latency evaluation, and error bars span the 25th to 75th percentiles across timed test batches. These interquartile ranges are very narrow and may therefore be obscured by the markers.}
\label{fig:budget_latency_main}
\end{figure}

Figure~\ref{fig:budget_latency_main} then maps these same inference settings to measured latency. LISTA is the fastest method, with one-stage LISTA around $0.03$ ms/sample in the measured setting. Adding Hybrid refinement raises latency to roughly $0.07$--$0.08$ ms/sample for one corrective step and about $0.33$--$0.37$ ms/sample for seven corrective steps. This is still far below the 100-step ISTA and MFISTA settings, which are several milliseconds per sample on the same plots. 

Together, Figs.~\ref{fig:pareto_main} and \ref{fig:budget_latency_main} show that the proposed method is not simply a compromise on an abstract budget axis. In measured wall-clock terms, Hybrid provides practical operating points between LISTA and fully iterative inference: it consistently outperforms the amortized baseline while remaining separated from long ISTA/MFISTA runs by a large latency margin.

\subsection{Hybrid performance across inference-budget allocations}
\label{sec:hybrid_allocation}

\begin{figure}[htbp]
\centering
\includegraphics[width=0.95\linewidth]{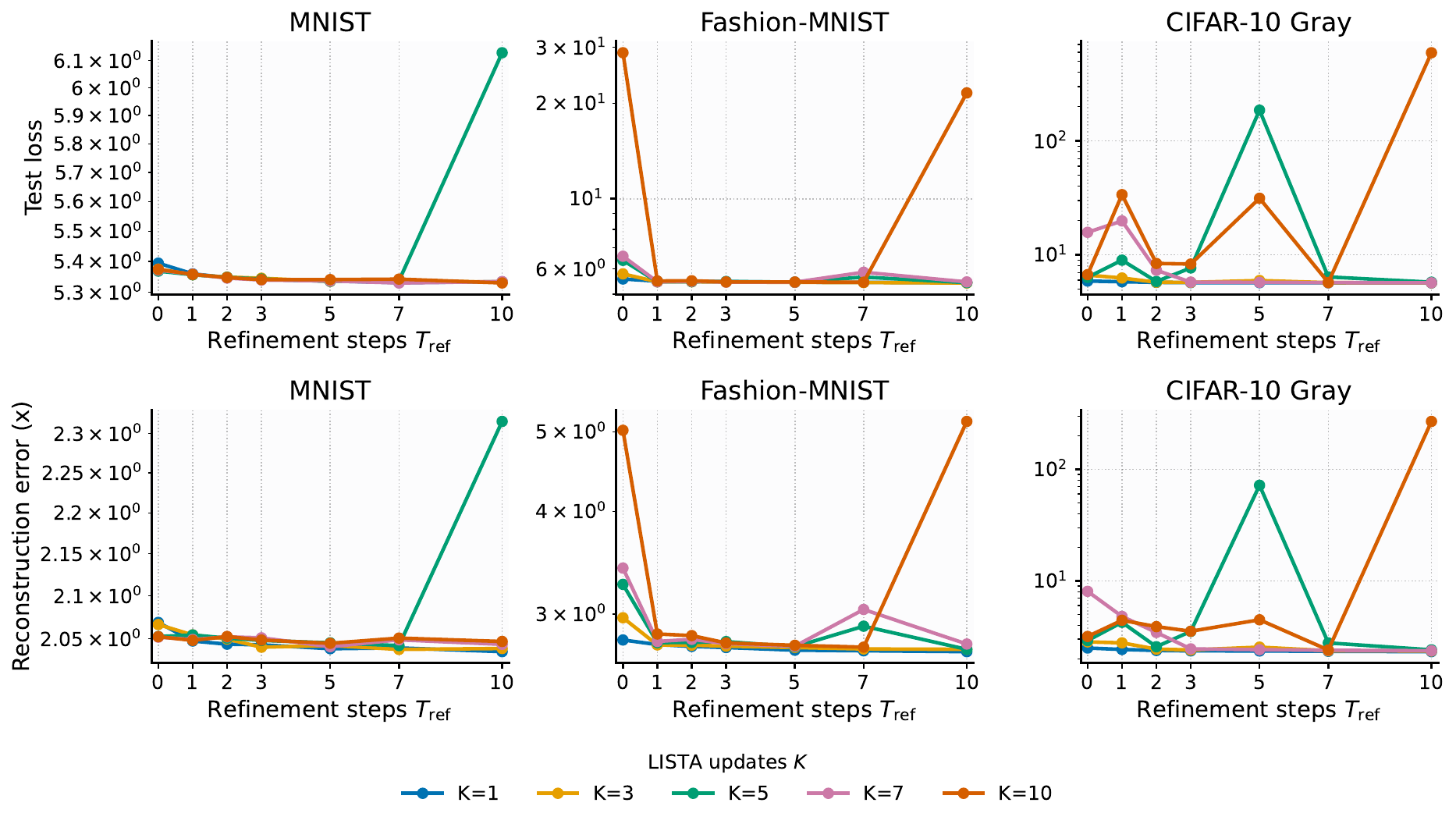}
\caption{Hybrid performance across inference-budget allocations on the main datasets. The horizontal axis shows the number of corrective refinement steps $T_{\mathrm{ref}}$, and different curves show different numbers of LISTA-style shrinkage stages $K$.}
\label{fig:hybrid_budget_main}
\end{figure}

Figure~\ref{fig:hybrid_budget_main} separates the two components of Hybrid inference: the number of LISTA-style amortized shrinkage stages $K$ inside the layerwise encoder and the number of corrective refinement steps $T_{\mathrm{ref}}$. The horizontal axis varies $T_{\mathrm{ref}}$, while different curves correspond to different values of $K$. Each $(K,T_{\mathrm{ref}})$ setting is trained separately.

Larger $K$ does not reliably yield better performance. On Fashion-MNIST and CIFAR-10 Gray, using more amortized stages without refinement ($T_{\mathrm{ref}}=0$) can perform worse than the $K=1$ configuration. By contrast, configurations combining $K=1$ with iterative refinement consistently outperform the corresponding pure-amortized procedure.

For $K=1$, most of the improvement relative to $T_{\mathrm{ref}}=0$ is achieved within the first few refinement steps on all three datasets and for both quality metrics, with smaller changes thereafter. This trend does not extend uniformly to larger $K$: performance can be non-monotonic, and some configurations deteriorate sharply. 

Table~\ref{tab:hybrid_budget_disentanglement} compares representative Hybrid configurations on Fashion-MNIST. Among settings with $K+T_{\mathrm{ref}}=6$, $(1,5)$ gives lower test loss and reconstruction error than $(5,1)$, whereas $(5,1)$ has lower latency; $(3,3)$ gives intermediate results. Thus, the two types of computation are not interchangeable: favoring refinement gives better quality in this comparison, whereas favoring amortized inference gives lower latency.

\begin{table}[htbp]
\centering
\caption{Representative Fashion-MNIST slice of the Hybrid inference-budget allocation sweep, restricted to settings with a matching dedicated latency measurement. Lower is better for test loss, reconstruction error, and latency. Quality values are means $\pm$ standard deviation across three runs with seeds $0$, $1$, and $2$. Latency values are medians with the 25th and 75th percentiles across timed test batches in brackets.}
\label{tab:hybrid_budget_disentanglement}
\small
\resizebox{\linewidth}{!}{
\begin{tabular}{cccccc}
\hline
$K$ & $T_{\mathrm{ref}}$ & $K+T_{\mathrm{ref}}$ & Test loss $\downarrow$ & Recon. error $\downarrow$ & ms/sample [P25, P75] $\downarrow$ \\
\hline
1 & 1 & 2 & $5.499 \pm 0.011$ & $2.754 \pm 0.004$ & $0.074$ [$0.074$, $0.075$] \\
1 & 3 & 4 & $5.479 \pm 0.011$ & $2.731 \pm 0.002$ & $0.172$ [$0.171$, $0.173$] \\
1 & 5 & 6 & $5.457 \pm 0.007$ & $2.711 \pm 0.004$ & $0.246$ [$0.246$, $0.247$] \\
1 & 7 & 8 & $5.449 \pm 0.006$ & $2.706 \pm 0.003$ & $0.333$ [$0.332$, $0.334$] \\
3 & 1 & 4 & $5.473 \pm 0.003$ & $2.751 \pm 0.002$ & $0.104$ [$0.104$, $0.105$] \\
3 & 3 & 6 & $5.461 \pm 0.001$ & $2.742 \pm 0.004$ & $0.191$ [$0.191$, $0.192$] \\
5 & 1 & 6 & $5.474 \pm 0.012$ & $2.769 \pm 0.004$ & $0.129$ [$0.129$, $0.129$] \\
5 & 3 & 8 & $5.483 \pm 0.027$ & $2.777 \pm 0.038$ & $0.220$ [$0.219$, $0.220$] \\
10 & 1 & 11 & $5.504 \pm 0.026$ & $2.837 \pm 0.052$ & $0.196$ [$0.196$, $0.197$] \\
10 & 3 & 13 & $5.467 \pm 0.004$ & $2.767 \pm 0.009$ & $0.283$ [$0.283$, $0.284$] \\
\hline
\end{tabular}
}
\end{table}

\subsection{Depth scaling}

\begin{figure}[htbp]
\centering
\includegraphics[width=0.95\linewidth]{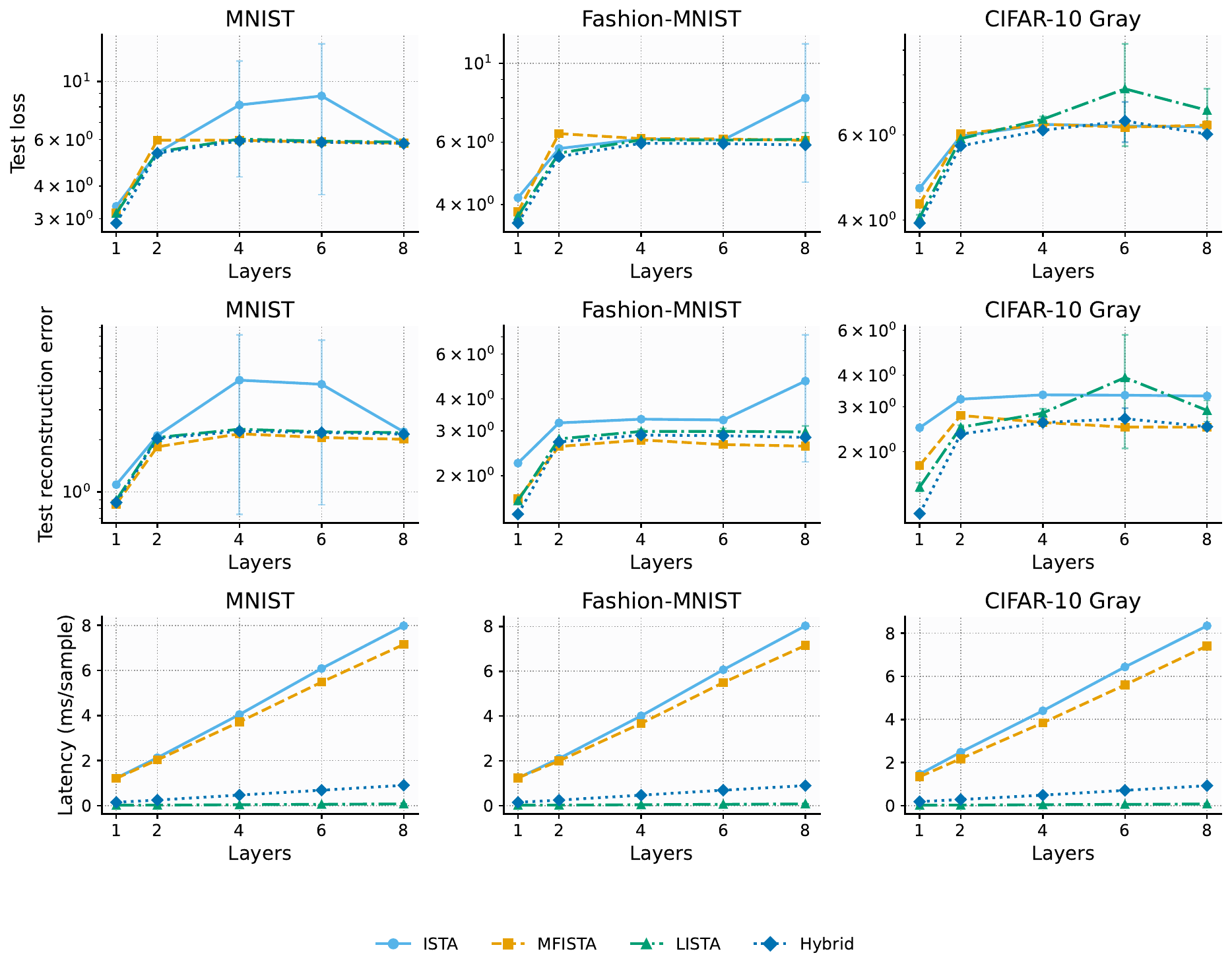}
\caption{Depth scaling across the main datasets. The panels show how objective quality and runtime change as the hierarchy deepens. Quality markers show means over three runs with seeds $0$, $1$, and $2$, and quality error bars indicate standard deviation across seeds. Latency markers show median ms/sample from the dedicated latency evaluation.}
\label{fig:depth_scaling_main}
\end{figure}

Figure~\ref{fig:depth_scaling_main} shows how quality and latency change as the hierarchy becomes deeper under the layer schedules specified in the experimental setup. The depth sweep shows that latency grows with depth, but the quality depends strongly on the inference method. MFISTA shows comparatively low run-to-run variability at most depths, whereas the ISTA-style baseline shows large loss variance in some regimes, including MNIST at depths 4 and 6. These patterns indicate that increasing hierarchy depth affects not only runtime but also the run-to-run variability of the learned procedures.

LISTA shows a different pattern. On Fashion-MNIST, increasing depth beyond the shallow settings does not clearly improve the final metrics. On CIFAR-10 Gray, LISTA remains fast but shows reduced quality at greater depths.

Hybrid remains much faster than the iterative solvers and usually shows lower run-to-run variability than pure LISTA and the ISTA-style baseline, although its results vary substantially across runs at some deeper CIFAR-10 Gray settings. Its refinement stage preserves competitive quality at most depths while avoiding the largest failures of the iterative baseline.

\subsection{Effect of sparsity strength and step-size scaling}

\begin{figure}[htbp]
\centering
\includegraphics[width=0.95\linewidth]{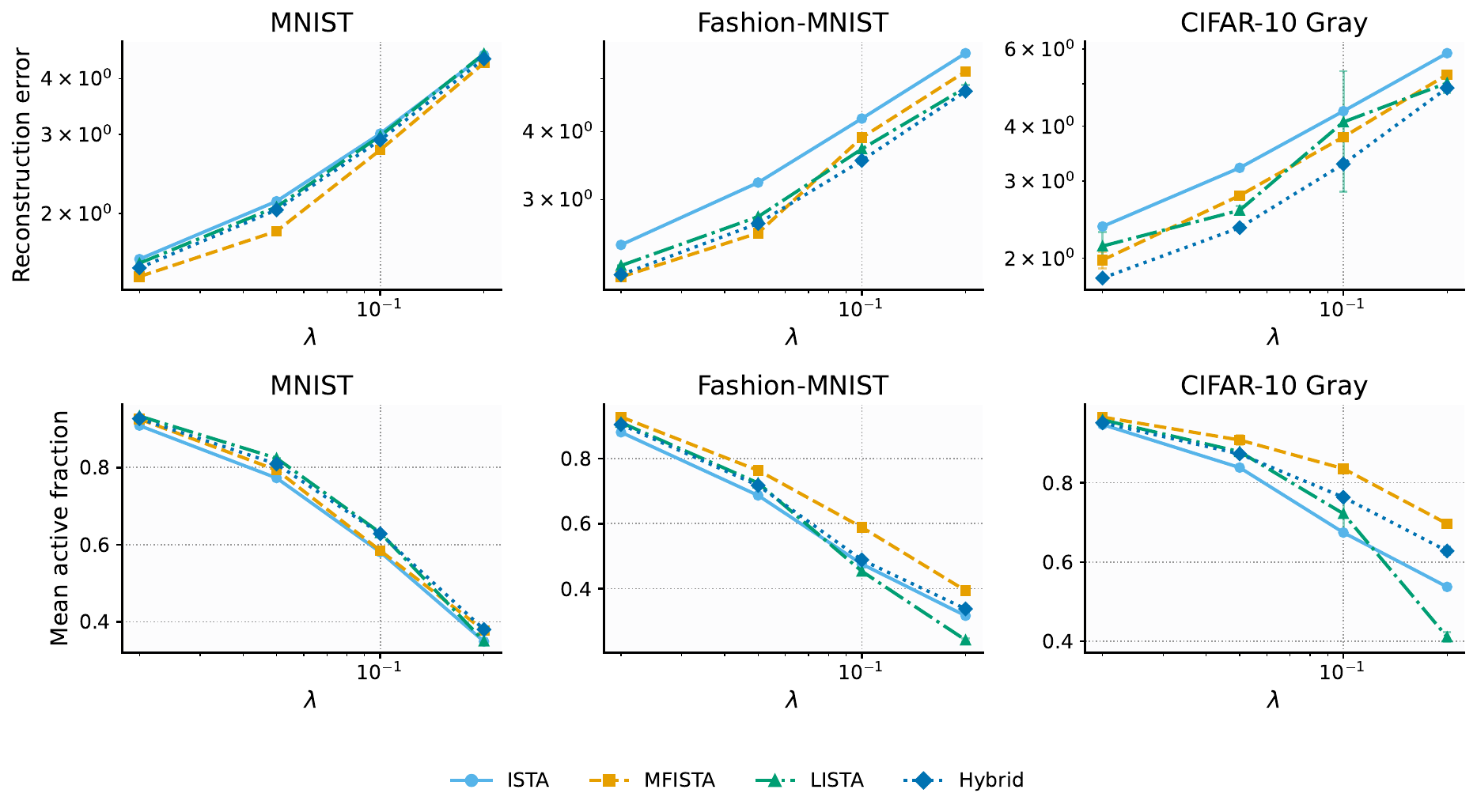}
\caption{Effect of sparsity strength on reconstruction error and mean active fraction. Markers show means over three runs with seeds $0$, $1$, and $2$, and error bars indicate standard deviation across seeds.}
\label{fig:sparsity_sweep}
\end{figure}

Figure~\ref{fig:sparsity_sweep} varies the tied sparsity coefficient $\lambda$ while keeping the rest of the default setup fixed. Increasing $\lambda$ reduces the mean active fraction and generally increases reconstruction error. Across the tested values, Hybrid generally achieves lower reconstruction error than pure LISTA, indicating that its advantage is not specific to the default value $\lambda=0.05$.

\begin{figure}[htbp]
\centering
\includegraphics[width=0.95\linewidth]{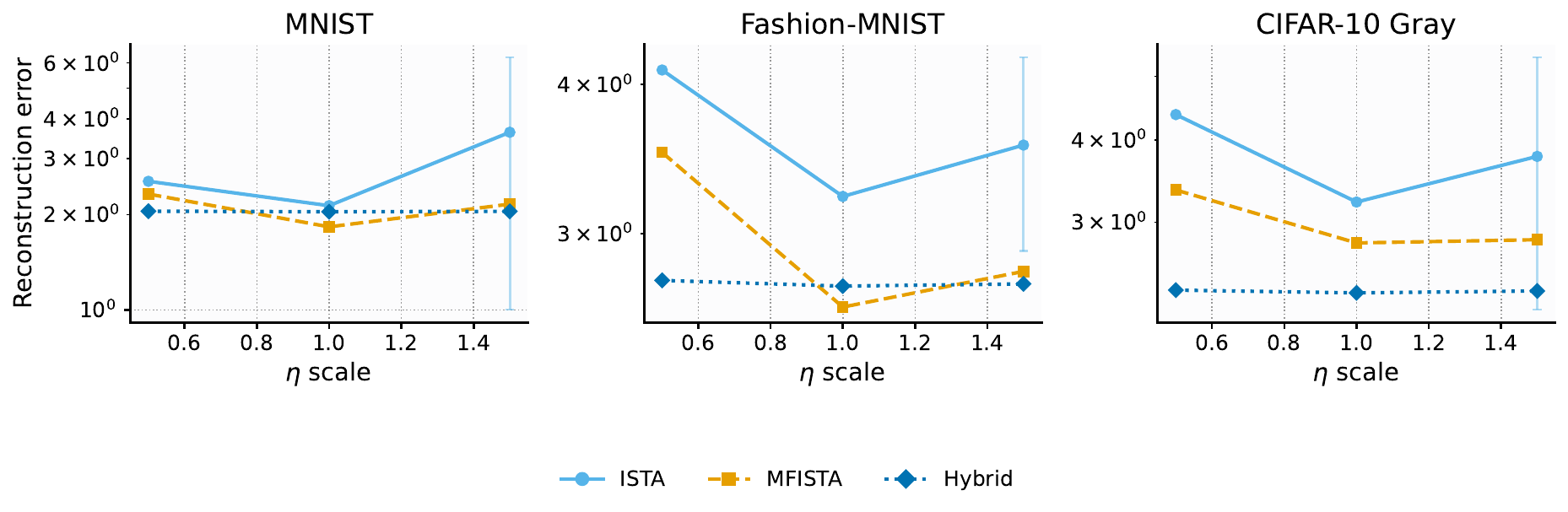}
\caption{Effect of step-size scaling on ISTA, MFISTA, and Hybrid. Markers show means over three runs with seeds $0$, $1$, and $2$, and error bars indicate standard deviation across seeds.}
\label{fig:ablation}
\end{figure}

Figure~\ref{fig:ablation} examines sensitivity to the refinement step-size scale by varying the same $\eta$-scale for ISTA, MFISTA, and Hybrid and comparing the resulting reconstruction error.

ISTA shows the clearest sensitivity to the step-size scale: $\eta_{\mathrm{scale}}=1.0$ gives the lowest mean reconstruction error across the plotted datasets, whereas increasing the scale to $1.5$ degrades its performance, particularly on MNIST. MFISTA varies more modestly, with $\eta_{\mathrm{scale}}=1.0$ giving either the lowest or a comparable reconstruction error. By contrast, Hybrid changes little across the tested scales.

\section{Discussion}

The experiments compare learned systems produced by different training-and-inference procedures, rather than inference algorithms operating on a single fixed dictionary. The observed differences therefore combine the direct effects of iterative or amortized inference with their indirect effects on dictionary learning and, for LISTA-based procedures, encoder learning. Within this practical end-to-end comparison, LISTA gives the lowest latency but can leave a quality gap on Fashion-MNIST, CIFAR-10 Gray, and deeper hierarchies. Procedures based on long iterative inference provide high-computation reference points. The Hybrid procedure occupies an intermediate operating regime, combining a small number of LISTA-style stages with a small number of objective-based corrective updates.

In the Hybrid budget-allocation experiment, increasing the number of LISTA-style stages is not a reliable substitute for refinement, whereas a few corrective steps after a shallow amortized pass frequently give a favorable quality--latency balance.

The LISTA-style encoder can provide a useful amortized initialization even with a single learned shrinkage stage, because it maps each input to a data-dependent sparse code. However, increasing the number of layerwise amortized stages is not equivalent to performing corrective inference under the full hierarchical energy. Additional shared LISTA-style stages repeatedly transform the bottom-up code within each layer, whereas Hybrid refinement explicitly uses reconstruction and inter-layer prediction errors from the coupled objective. This explains why $K=1$ can be competitive, while larger $K$ does not reliably replace iterative refinement.

The depth, sparsity, and step-size checks clarify the scope of this pattern. Depth scaling shows that Hybrid can reduce run-to-run variability relative to pure LISTA and the unaccelerated ISTA-style baseline, but this reduction does not hold at every depth. The sparsity sweep shows that the qualitative behavior is not tied to one value of $\lambda$. The step-size ablation shows that Hybrid changes little across the tested scales, while the iterative baselines are more sensitive.

The present experiments leave two important limitations. The implementation does not constitute a fully local biological learning rule, and the observed run-to-run variability of the trained Hybrid procedures does not establish a general convergence guarantee. Establishing more biologically local learning mechanisms and formalizing the convergence properties of the resulting inference dynamics remain important directions for future work.

Future work should test whether this hybrid principle scales beyond the fully connected static-image setting studied here. One direction is convolutional sparse coding and larger image models, where dictionary structure and spatial weight sharing may change both reconstruction quality and inference cost \citep{Bristow:2013,Boutin:2021}. Another direction is to combine the present framework with more advanced components from modern sparse-coding and algorithm-unrolling methods, such as structured amortized encoders, ALISTA-style parameterizations \citep{Liu:2019}, adaptive inference budgets, or safeguarded refinement schemes. These extensions may improve the quality--latency trade-off in more demanding domains without changing the central idea of using amortized inference as a fast initializer for sparse iterative correction.

Finally, extending the framework to temporal data would test whether Hybrid inference remains effective for continuous inputs under tighter latency constraints.

\section{Conclusion}

We compared hierarchical sparse-coding procedures that share the same objective formulation and architecture but incorporate iterative, LISTA-style amortized, or Hybrid latent inference. Across the static-image benchmarks, pure amortization achieved the lowest latency but sometimes showed gaps in reconstruction quality and greater run-to-run variability, whereas long iterative inference incurred substantially greater cost. Hybrid configurations with a shallow amortized pass followed by a few refinement steps frequently provided a favorable quality--latency balance. Overall, the results support Hybrid inference as a practical component of hierarchical sparse coding.

\section*{Acknowledgements}

The author used coding agents, including Codex, OpenCode and GitHub Copilot, with GPT-5.6, GPT-5.5, GPT-5.4, Qwen 3.6 27B, and Qwen 3.6 35B to help revise wording, improve clarity, check the consistency of the manuscript, and write and revise program code. The scientific ideas, experimental design, implementation, analysis, interpretation, and final text were reviewed and approved by the author, who takes full responsibility for the content of the paper.

%\bibliography{sparse}

\begin{thebibliography}{}

\bibitem[Abe et~al., 2018]{Abe:2018}
Abe, Y., Fujita, K., and Kashimori, Y. (2018).
\newblock Visual and category representations shaped by the interaction between
  inferior temporal and prefrontal cortices.
\newblock {\em Cognitive Computation}, 10:687 -- 702.

\bibitem[Aberdam et~al., 2019]{Aberdam:2019}
Aberdam, A., Sulam, J., and Elad, M. (2019).
\newblock Multi-layer sparse coding: The holistic way.
\newblock {\em SIAM Journal on Mathematics of Data Science}, 1(1):46--77.

\bibitem[Ablin et~al., 2019]{Ablin:2019}
Ablin, P., Moreau, T., Massias, M., and Gramfort, A. (2019).
\newblock Learning step sizes for unfolded sparse coding.
\newblock In {\em Advances in Neural Information Processing Systems 32 (NeurIPS
  2019)}, pages 13100--13110.

\bibitem[Arbelaez et~al., 2011]{Arbelaez:2011}
Arbelaez, P., Maire, M., Fowlkes, C., and Malik, J. (2011).
\newblock Contour detection and hierarchical image segmentation.
\newblock {\em IEEE Trans. Pattern Anal. Mach. Intell.}, 33(5):898--916.

\bibitem[Bastos et~al., 2012]{Bastos:2012}
Bastos, A., Usrey, W., Adams, R., Mangun, G., Fries, P., and Friston, K.
  (2012).
\newblock Canonical microcircuits for predictive coding.
\newblock {\em Neuron}, 76(4):695--711.

\bibitem[Beck and Teboulle, 2009a]{Beck:2009a}
Beck, A. and Teboulle, M. (2009a).
\newblock Fast gradient-based algorithms for constrained total variation image
  denoising and deblurring problems.
\newblock {\em IEEE Transactions on Image Processing}, 18(11):2419--2434.

\bibitem[Beck and Teboulle, 2009b]{Beck:2009b}
Beck, A. and Teboulle, M. (2009b).
\newblock A fast iterative shrinkage-thresholding algorithm for linear inverse
  problems.
\newblock {\em SIAM Journal on Imaging Sciences}, 2(1):183--202.

\bibitem[Beck and Tetruashvili, 2013]{Beck:2013}
Beck, A. and Tetruashvili, L. (2013).
\newblock On the convergence of block coordinate descent type methods.
\newblock {\em SIAM Journal on Optimization}, 23(4):2037--2060.

\bibitem[Boutin et~al., 2021]{Boutin:2021}
Boutin, V., Franciosini, A., Chavane, F., Ruffier, F., and Perrinet, L. (2021).
\newblock Sparse deep predictive coding captures contour integration
  capabilities of the early visual system.
\newblock {\em PLOS Computational Biology}, 17(1):1--31.

\bibitem[Bristow et~al., 2013]{Bristow:2013}
Bristow, H., Eriksson, A., and Lucey, S. (2013).
\newblock Fast convolutional sparse coding.
\newblock In {\em Proceedings of the IEEE Conference on Computer Vision and
  Pattern Recognition (CVPR)}, pages 391--398.

\bibitem[Daubechies et~al., 2004]{Daubechies:2004}
Daubechies, I., Defrise, M., and De~Mol, C. (2004).
\newblock An iterative thresholding algorithm for linear inverse problems with
  a sparsity constraint.
\newblock {\em Communications on Pure and Applied Mathematics},
  57(11):1413--1457.

\bibitem[Demmel, 1997]{Demmel:1997}
Demmel, J.~W. (1997).
\newblock {\em Applied Numerical Linear Algebra}.
\newblock Society for Industrial and Applied Mathematics.

\bibitem[Friston, 2005]{Friston:2005}
Friston, K.~J. (2005).
\newblock A theory of cortical responses.
\newblock {\em Philosophical Transactions of the Royal Society B: Biological
  Sciences}, 360:815 -- 836.

\bibitem[Friston and Kiebel, 2009]{Friston:2009}
Friston, K.~J. and Kiebel, S.~J. (2009).
\newblock Predictive coding under the free-energy principle.
\newblock {\em Philosophical Transactions of the Royal Society B: Biological
  Sciences}, 364:1211 -- 1221.

\bibitem[Golub and van Loan, 2013]{Golub:2013}
Golub, G.~H. and van Loan, C.~F. (2013).
\newblock {\em Matrix Computations}.
\newblock JHU Press, fourth edition.

\bibitem[Gregor and LeCun, 2010]{Gregor:2010}
Gregor, K. and LeCun, Y. (2010).
\newblock Learning fast approximations of sparse coding.
\newblock In {\em Proceedings of the 27th International Conference on
  International Conference on Machine Learning}, ICML'10, page 399–406,
  Madison, WI, USA. Omnipress.

\bibitem[Kamiyama et~al., 2016]{Kamiyama:2016}
Kamiyama, A., Fujita, K., and Kashimori, Y. (2016).
\newblock A neural mechanism of dynamic gating of task-relevant information by
  top-down influence in primary visual cortex.
\newblock {\em Biosystems}, 150:138--148.

\bibitem[Kashimori et~al., 2007]{Kashimori:2007}
Kashimori, Y., Ichinose, Y., and Fujita, K. (2007).
\newblock A functional role of interaction between it cortex and pf cortex in
  visual categorization task.
\newblock {\em Neurocomputing}, 70(10):1813--1818.
\newblock Computational Neuroscience: Trends in Research 2007.

\bibitem[Liu et~al., 2019]{Liu:2019}
Liu, J., Chen, X., Wang, Z., and Yin, W. (2019).
\newblock Alista: Analytic weights are as good as learned weights in lista.
\newblock In {\em International Conference on Learning Representations}.

\bibitem[Mairal et~al., 2010]{Mairal:2010}
Mairal, J., Bach, F., Ponce, J., and Sapiro, G. (2010).
\newblock Online learning for matrix factorization and sparse coding.
\newblock {\em J. Mach. Learn. Res.}, 11:19--60.

\bibitem[Nesterov, 2012]{Nesterov:2012}
Nesterov, Y. (2012).
\newblock Efficiency of coordinate descent methods on huge-scale optimization
  problems.
\newblock {\em SIAM Journal on Optimization}, 22(2):341--362.

\bibitem[Olshausen and Field, 1996]{Olshausen:1996}
Olshausen, B.~A. and Field, D.~J. (1996).
\newblock Emergence of simple-cell receptive field properties by learning a
  sparse code for natural images.
\newblock {\em Nature}, 381:607--609.

\bibitem[Parikh and Boyd, 2014]{Parikh:2014}
Parikh, N. and Boyd, S. (2014).
\newblock Proximal algorithms.
\newblock {\em Found. Trends Optim.}, 1(3):127–239.

\bibitem[Rao and Ballard, 1999]{Rao:1999}
Rao, R. and Ballard, D. (1999).
\newblock Predictive coding in the visual cortex: a functional interpretation
  of some extra-classical receptive-field effects.
\newblock {\em Nature neuroscience}, 2:79--87.

\bibitem[Richt\'{a}rik and Tak\'{a}\u{c}, 2014]{Richtarik:2014}
Richt\'{a}rik, P. and Tak\'{a}\u{c}, M. (2014).
\newblock Iteration complexity of randomized block-coordinate descent methods
  for minimizing a composite function.
\newblock {\em Math. Program.}, 144(1–2):1–38.

\bibitem[Rosenbaum, 2022]{Rosenbaum:2022}
Rosenbaum, R. (2022).
\newblock On the relationship between predictive coding and backpropagation.
\newblock {\em PLOS ONE}, 17(3):1--27.

\bibitem[Tschantz et~al., 2023]{Tschantz:2023}
Tschantz, A., Millidge, B., Seth, A.~K., and Buckley, C.~L. (2023).
\newblock Hybrid predictive coding: Inferring, fast and slow.
\newblock {\em PLOS Computational Biology}, 19(8):1--31.

\bibitem[Whittington and Bogacz, 2017]{Whittington:2017}
Whittington, J. C.~R. and Bogacz, R. (2017).
\newblock An approximation of the error backpropagation algorithm in a
  predictive coding network with local hebbian synaptic plasticity.
\newblock {\em Neural Computation}, 29(5):1229--1262.

\bibitem[Xu, 2018]{Xu:2018}
Xu, Y. (2018).
\newblock Hybrid jacobian and gauss--seidel proximal block coordinate update
  methods for linearly constrained convex programming.
\newblock {\em SIAM Journal on Optimization}, 28(1):646--670.

\end{thebibliography}

\appendix
\section{MFISTA-style accelerated inference}
\label{app:mfista}

For empirical comparison, we introduce an MFISTA-based iterative baseline evaluated under the same energy and inference-budget protocol. MFISTA, following the monotone accelerated scheme of \citet{Beck:2009a}, is included only as an accelerated iterative reference. It uses the same hierarchical sparse energy and the same proximal shrinkage operation as the ISTA-style baseline, but differs in two implementation details. First, the gradient update is computed at an extrapolated point rather than at the current iterate. Second, the resulting joint latent-code candidate is accepted only if it lowers the full energy. This acceptance rule applies to the inferred codes, not to the dictionary parameters.

Let $\bar{\ba}^{(t)} = (\bar{\ba}_1^{(t)}, \dots, \bar{\ba}_L^{(t)})$ denote the accepted latent codes at iteration $t$, let $\bm{y}^{(t)} = (\bm{y}_1^{(t)}, \dots, \bm{y}_L^{(t)})$ denote the extrapolated codes, and let $s_t$ be the FISTA momentum scalar. We initialize the momentum as $s_0=1$ and set the first extrapolated code equal to the initial accepted code, $\bm{y}^{(0)}=\bar{\ba}^{(0)}$. We first compute the same layerwise ISTA step-size estimates $(\{\eta_\ell\}_{\ell=1}^{L})$ as in Eq.~\eqref{eq:stepsize_rule}, using the current dictionaries, coupling weights, and $\eta_{\mathrm{scale}}$. For the accelerated baseline, we then collapse these layerwise estimates to a single fixed step size shared across all latent blocks:
\begin{equation}
\eta = \min_{\ell} \eta_\ell,
\qquad
\theta_\ell = \eta \lambda_\ell.
\end{equation}
This choice is conservative for the layerwise ISTA updates, but does not guarantee a globally valid step size for the coupled objective.

At each iteration, MFISTA forms the same kind of soft-thresholded proximal candidate as ISTA, but at the extrapolated point:
\begin{equation}
\bm{z}_\ell^{(t)}
=
\mathcal{S}_{\theta_\ell }
\left(
\bm{y}_\ell^{(t)}
-
\eta
\nabla_{\ba_\ell} f(\bm{y}^{(t)})
\right),
\qquad
\ell=1,\ldots,L .
\end{equation}
Here, $\nabla_{\ba_\ell} f(\bm{y}^{(t)})$ denotes the $\ell$-th block gradient of the same smooth energy used in ISTA, evaluated at the full extrapolated code $\bm{y}^{(t)}$.
Equivalently, it is obtained from the appropriate layerwise expression in Eqs.~\eqref{eq:grad_a1}--\eqref{eq:grad_aL}, according to whether $\ell=1$, $1<\ell<L$, or $\ell=L$, after substituting $\ba_j=\bm{y}_j^{(t)}$ for all latent blocks. The full candidate vector $\bm{z}^{(t)} = (\bm{z}_1^{(t)}, \dots, \bm{z}_L^{(t)})$ is then compared with the previously accepted joint latent code:
\begin{equation}
\bar{\ba}^{(t+1)} = \arg\min_{\bm{u} \in \{\bar{\ba}^{(t)}, \bm{z}^{(t)}\}} E(\bm{u}),
\end{equation}
so the accepted energy is monotone non-increasing over MFISTA inference iterations. The momentum scalar is updated by
\begin{equation}
s_{t+1}
=
\frac{1+\sqrt{1+4s_t^2}}{2}.
\end{equation}
The next extrapolated latent code is formed layerwise via
\begin{equation}
\bm{y}_\ell^{(t+1)}
=
\bar{\ba}_\ell^{(t+1)}
+
\frac{s_t}{s_{t+1}}\left(\bm{z}_\ell^{(t)}-\bar{\ba}_\ell^{(t+1)}\right)
+
\frac{s_t-1}{s_{t+1}}\left(\bar{\ba}_\ell^{(t+1)}-\bar{\ba}_\ell^{(t)}\right),
\qquad \ell=1,\dots,L.
\end{equation}

\section{Supplementary Results for the Main Experiments}
\label{app:supplementary_results}

This appendix section supplements the main experiments with training-time convergence curves and an additional ablation of the Hybrid refinement rule. These analyses use the same objective, architecture family, and training protocol as the main text. The separate appendix-only BSDS500 Patch experiment is presented in Section~\ref{app:natural_image_results}.

\subsection{Training-time convergence in epoch space}
\label{app:e1_convergence}

\begin{figure}[htbp]
\centering
\includegraphics[width=0.95\linewidth]{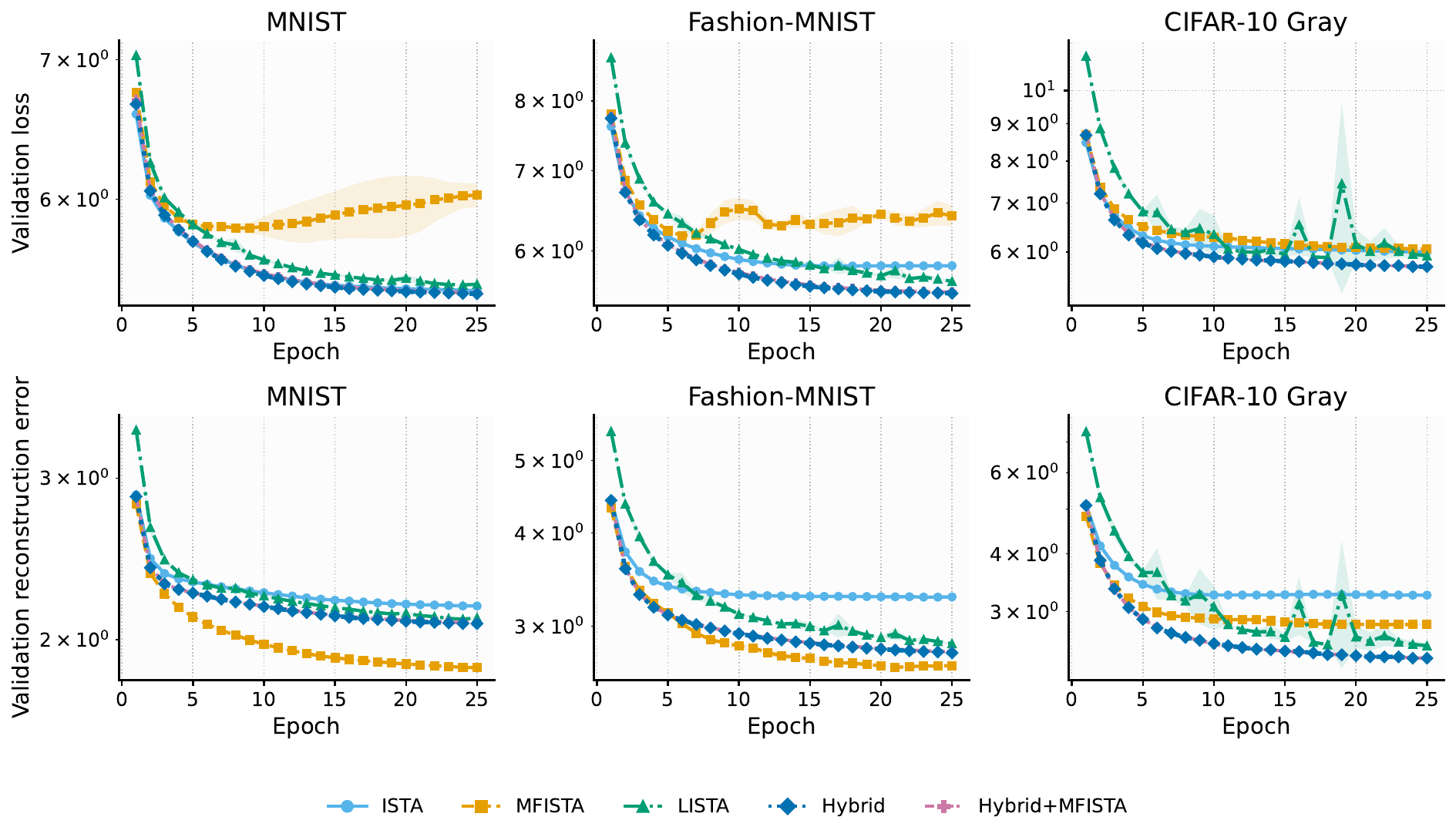}
\caption{Training-time convergence curves across optimization methods. The panels show validation loss and reconstruction error as functions of training epoch under the shared core setting. Validation loss is the full hierarchical sparse energy, whereas reconstruction error denotes only the input-space reconstruction term. Curves show means over three runs with seeds $0$, $1$, and $2$, and shaded bands indicate standard deviation across seeds.}
\label{fig:e1_convergence}
\end{figure}

Figure~\ref{fig:e1_convergence} shows the training-time evolution of validation loss and reconstruction error over the 25-epoch schedule used in the main experiments. This check assesses whether the shared schedule is sufficient for the reported comparisons.

The curves show that all methods improve rapidly in the early epochs and are comparatively flat by the end of training. Thus, the 25-epoch schedule appears adequate for comparing methods under this fixed protocol. The remaining separation among methods is already visible late in training: LISTA does not close the quality gap entirely, while Hybrid moves closer to the iterative reference regime after refinement.

\subsection{Additional ablation: Hybrid-MFISTA refinement}
\label{app:hybrid_mfista}

The proposed Hybrid method uses ISTA-style refinement. As an additional ablation, we also tested Hybrid-MFISTA, which replaces only the refinement stage with the MFISTA procedure in Appendix~\ref{app:mfista}. The amortized initialization, encoder-gradient convention, and training protocol are unchanged.

In the full set of method-wise epoch curves in Fig.~\ref{fig:e1_convergence}, the Hybrid-MFISTA curves track the original Hybrid method closely over training epochs (the two curves overlap almost completely). Table~\ref{tab:hybrid_mfista_supp} shows that Hybrid-MFISTA is sometimes competitive but does not consistently outperform Hybrid. MFISTA-style refinement adds latency without a clear objective-quality gain. This ablation supports using the simpler ISTA-style refinement rule in the main Hybrid method.

\begin{table}[htbp]
\centering
\caption{Appendix additional ablation comparing the main Hybrid method with Hybrid-MFISTA refinement. Quality values are means $\pm$ standard deviation across three runs with seeds $0$, $1$, and $2$. Latency values are medians with the 25th and 75th percentiles across timed test batches in brackets.}
\label{tab:hybrid_mfista_supp}
\resizebox{0.88\linewidth}{!}{
\begin{tabular}{lcccc}
\hline
Dataset & Method & Test loss $\downarrow$ & Recon. error $\downarrow$ & ms/sample [P25, P75] $\downarrow$ \\
\hline
MNIST & Hybrid & $5.336 \pm 0.000$ & $2.039 \pm 0.003$ & $0.248$ [$0.247$, $0.249$] \\
MNIST & Hybrid-MFISTA & $5.342 \pm 0.004$ & $2.042 \pm 0.005$ & $0.563$ [$0.562$, $0.564$] \\
Fashion-MNIST & Hybrid & $5.457 \pm 0.007$ & $2.711 \pm 0.004$ & $0.246$ [$0.246$, $0.247$] \\
Fashion-MNIST & Hybrid-MFISTA & $5.467 \pm 0.010$ & $2.717 \pm 0.003$ & $0.577$ [$0.576$, $0.578$] \\
CIFAR-10 Gray & Hybrid & $5.694 \pm 0.010$ & $2.341 \pm 0.011$ & $0.286$ [$0.286$, $0.287$] \\
CIFAR-10 Gray & Hybrid-MFISTA & $5.699 \pm 0.009$ & $2.346 \pm 0.008$ & $0.621$ [$0.620$, $0.622$] \\
\hline
\end{tabular}
}
\end{table}

\section{Appendix-only dataset experiments}
\label{app:natural_image_results}

We additionally evaluate BSDS500 Patch to connect the comparison to the classical natural-image sparse-coding setting. The implementation samples $16 \times 16$ patches from the official BSDS500 train, validation, and test image directories, using 54000, 6000, and 10000 patches for the three splits, respectively.

\begin{figure}[htbp]
\centering
\includegraphics[width=0.95\linewidth]{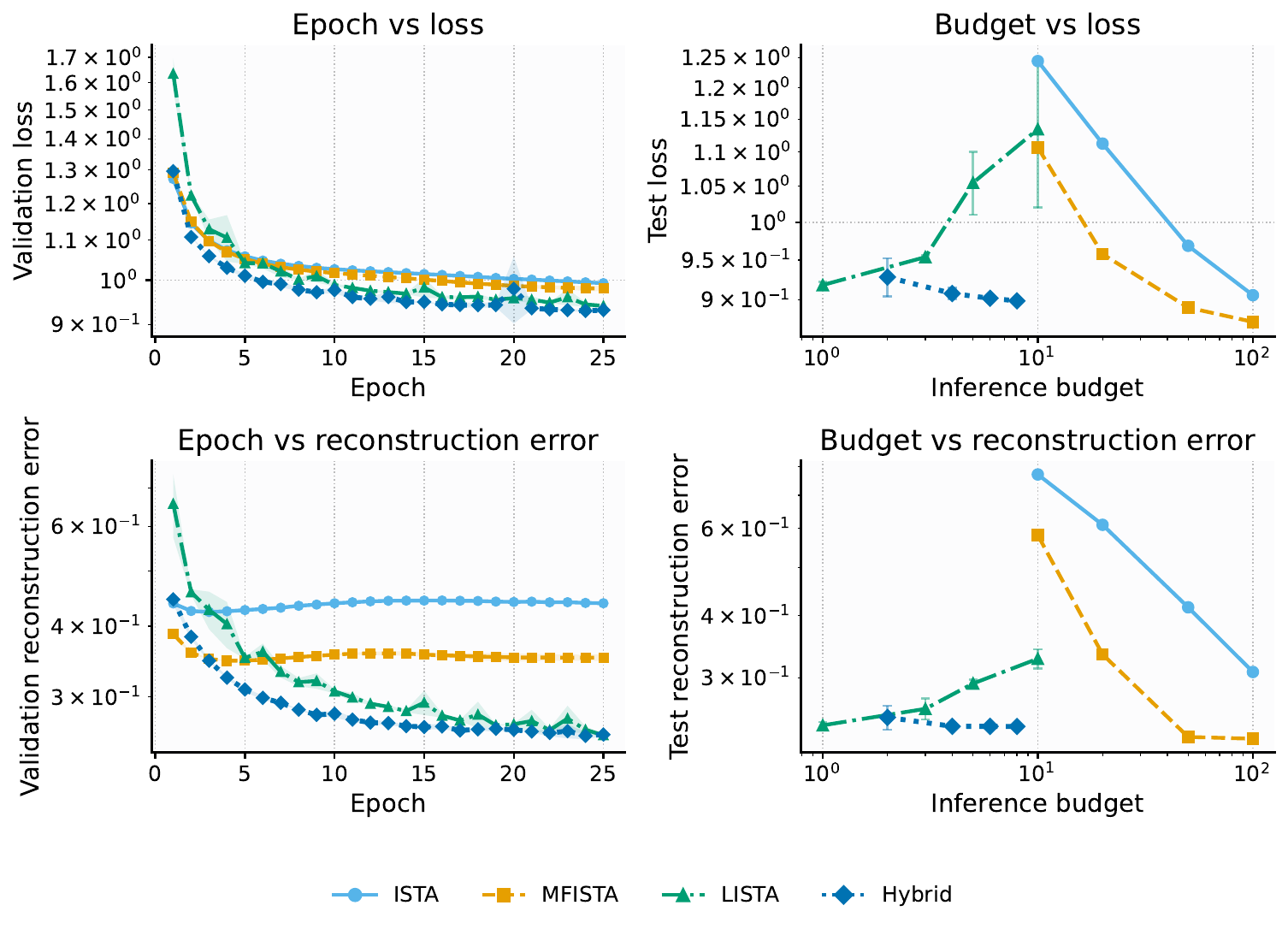}
\caption{BSDS500 Patch appendix check. Left column: validation loss and reconstruction error over training epochs. Right column: test loss and reconstruction error versus inference budget. For Hybrid in the budget panels, the amortized depth is fixed at $K=1$, matching the main quality-versus-budget comparison. Curves and markers show means over three runs with seeds $0$, $1$, and $2$; shaded bands in the left column and error bars in the right column indicate standard deviation across seeds.}
\label{fig:bsds500_appendix_summary}
\end{figure}

Figure~\ref{fig:bsds500_appendix_summary} checks two points. First, the epoch curves rule out an obvious training-schedule mismatch: the methods improve rapidly and flatten within the common 25-epoch schedule. Second, the budget panels ask whether the main quality-versus-inference-budget comparison remains recognizable on natural-image patches. The BSDS500 Patch result is more nuanced than the main grayscale benchmarks, but the same broad trade-off remains visible: pure amortization is cheapest, stronger iterative inference can improve final quality, and short Hybrid refinement occupies an intermediate operating regime.

\end{document}